\documentclass[journal,twoside,web]{ieeecolor}
\usepackage{jsen}
\usepackage{cite}
\usepackage{amsmath,amssymb,amsfonts}
\usepackage{algorithmic}
\usepackage{graphicx}
\usepackage{textcomp}
\usepackage{wrapfig}
\usepackage{multirow}
\usepackage{booktabs}
\usepackage{makecell}
\usepackage{tabularx}

\usepackage{subcaption}

\usepackage{array}
\usepackage{url}

\graphicspath{{figures/}}
\DeclareGraphicsExtensions{.pdf,.png,.jpg,.jpeg}

\def\BibTeX{{\rm B\kern-.05em{\sc i\kern-.025em b}\kern-.08em
    T\kern-.1667em\lower.7ex\hbox{E}\kern-.125emX}}
\definecolor{abstractbg}{rgb}{0.89804,0.94510,0.83137}
\begin{document}
\title{DPGS-SLAM: Diffusion-Prior and Provenance-Aware Monocular Gaussian Splatting SLAM for Dynamic Visual Sensing}
\author{Yifan Luo, Takafumi Matsumaru, \IEEEmembership{Senior Member, IEEE}, Vibekananda Dutta, \IEEEmembership{Member, IEEE}, and Xin He, \IEEEmembership{Member, IEEE}
\thanks{This work was supported in part by Waseda University through its Grant for Special Research Projects (Project number: 2025Q-026, 2025C-524, 2026C-529). The work was also supported in part by the NCN SONATA program, under Grant UMO-2024/55/D/ST7/02627, Poland.}
\thanks{Yifan Luo and Takafumi Matsumaru are with the Graduate School of Information, Production and Systems, Waseda University, Kitakyushu, Fukuoka 808-0135, Japan (e-mail: luoyifan01@ruri.waseda.jp; matsumarum@waseda.jp).}
\thanks{Vibekananda Dutta and Xin He are with the Faculty of Mechatronics, Warsaw University of Technology, 00-661 Warsaw, Poland (e-mail: vibekananda.dutta@pw.edu.pl; xin.he@pw.edu.pl).}
\thanks{Vibekananda Dutta is also with the School of Mechanical Engineering, Shanghai Jiao Tong University, Shanghai 200240, China.}%
}

\IEEEtitleabstractindextext{%
\fcolorbox{abstractbg}{abstractbg}{%
\begin{minipage}{\textwidth}%
\begin{wrapfigure}[15]{r}{4in}%
\includegraphics[width=4in]{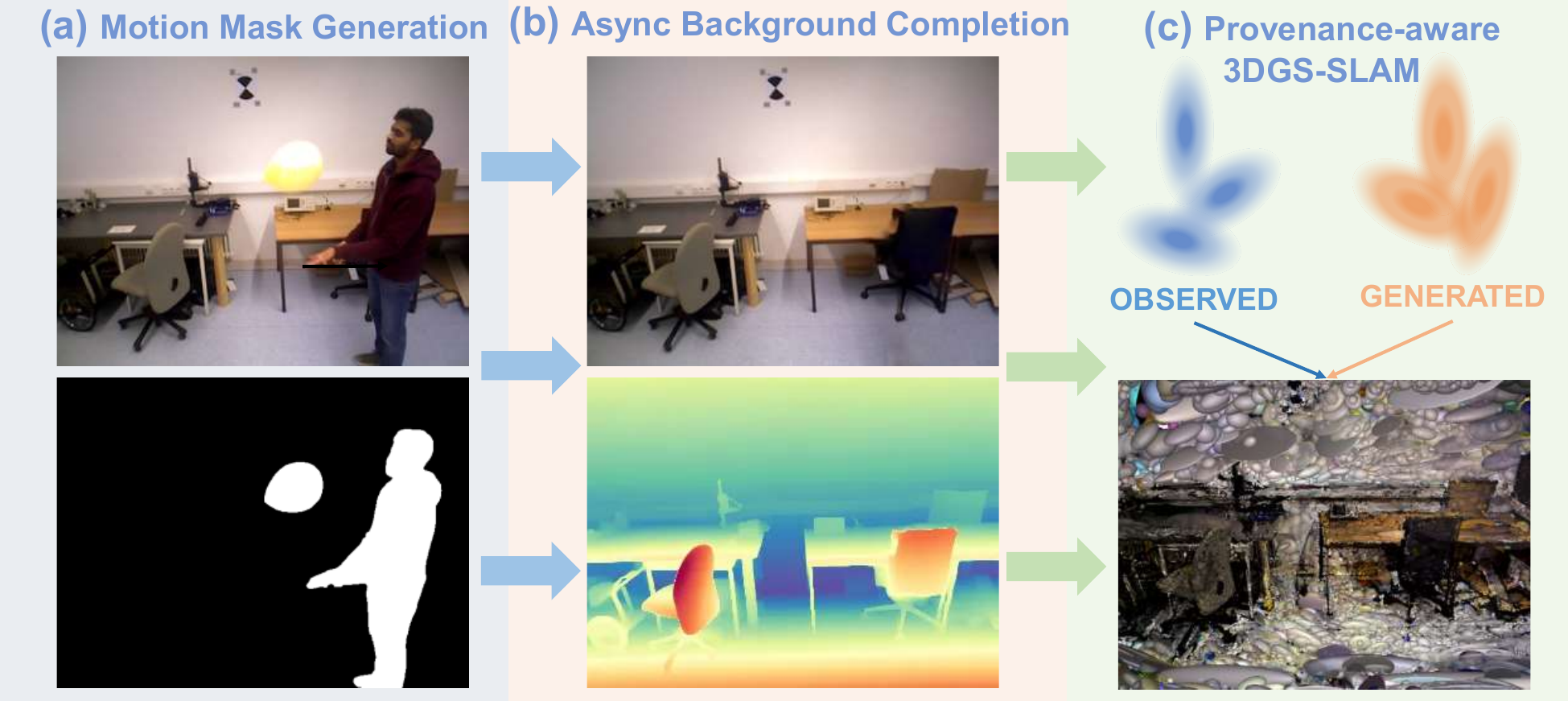}%
\end{wrapfigure}%
\begin{abstract}
This paper presents DPGS-SLAM (Diffusion-Prior and Provenance-Aware Monocular Gaussian Splatting SLAM) for dense mapping in environments containing dynamic objects.
Objects in dynamic motion remain a significant challenge for dense visual SLAM (simultaneous localization and mapping) since dynamic observations can contaminate camera tracking and scene reconstruction. Although 3DGS (3D Gaussian splatting) enables efficient dense mapping, existing 3DGS-SLAM systems may exhibit instability when dynamic objects violate their underlying static-scene assumption. Excluding dynamic observations can reduce their influence on tracking and mapping, but may leave occluded static backgrounds missing or weakly constrained.
Our proposed DPGS-SLAM is a monocular RGB SLAM framework that supports provisional representations of insufficiently observed static backgrounds through asynchronous diffusion-based background completion and provenance-aware Gaussian mapping. Dynamic regions are identified through optical-flow-based motion discovery and temporally consistent mask propagation by SAM2 (segment anything model 2). For selected keyframes, a fine-tuned diffusion inpainting model generates plausible background appearance in occluded regions, while metric depth predicted from completed images initializes provisional Gaussian primitives. Provenance labels distinguish generated primitives from observation-supported primitives, excluding generated content from camera-pose estimation and observed-map optimization. When previously occluded backgrounds become visible, eligible generated primitives are removed, and observation-supported primitives are independently initialized from real RGB observations and predicted depth.
Experimental results demonstrated competitive camera tracking, with average ATE RMSE (absolute trajectory error, reported as root mean squared error) values of 1.2\,cm on TUM RGB-D and 2.7\,cm on BONN RGB-D datasets. On StaDy4D, DPGS-SLAM improved average PSNR (peak signal-to-noise ratio) by 4.60\,dB over WildGS-SLAM using only monocular RGB input.
\end{abstract}

\end{minipage}}}

\maketitle

\section{Introduction}
\label{sec:introduction}

\IEEEPARstart{T}{his} paper proposes \textbf{DPGS-SLAM} (Diffusion-Prior and Provenance-Aware Monocular Gaussian Splatting SLAM), a monocular RGB SLAM (simultaneous localization and mapping) framework for dynamic visual sensing, designed to address insufficient real observation support for static-background reconstruction through asynchronous static-background completion via diffusion-based inpainting~\cite{diffusion,latentDiffusion} and provenance-aware Gaussian mapping. 
Here, \textbf{inpainting} refers to the general technique of synthesizing image content within masked regions, whereas \textbf{background completion} refers to the specific task of generating plausible static-background content in regions occluded by dynamic objects. A real observation of a static-background surface is defined as an RGB measurement acquired when that surface is directly visible to the camera. Insufficient real observation support occurs when such measurements are absent or too sparsely distributed across the available views to reliably constrain the reconstruction of the corresponding surface.

Achieving reliable SLAM in dynamic environments remains a long-standing challenge in the computer vision community. Many traditional SLAM systems, such as ORB-SLAM (2015)~\cite{orbslam} and DSO (2017)~\cite{dso}, are built upon the assumption of a rigid scene and therefore tend to suffer from tracking drift or reconstruction instability when independently moving objects are present.

Recently, \textbf{NeRF} (neural radiance fields) (2020)~\cite{nerf} -based volumetric representations have been explored as a dense-mapping paradigm for SLAM, due to their strong photometric modeling capability and high-quality novel-view synthesis. 
However, NeRF-SLAM methods typically incur high computational costs and limited rendering efficiency, and often rely on additional depth sensors or predefined dynamic priors, such as DN-SLAM (2023)~\cite{dnslam} and RoDyn-SLAM (2024)~\cite{rodyn}, to handle dynamic regions. 
These constraints restrict their applicability in purely monocular RGB settings and limit their robustness in complex dynamic scenes.

More recently, \textbf{3DGS} (3D Gaussian splatting) (2023)~\cite{3dgs} has emerged as an efficient and expressive scene representation for dense scene modeling. By modeling scenes as a collection of learnable 3D Gaussian primitives and enabling fast, differentiable rasterization, 3DGS has been widely adopted in recent SLAM systems, such as SplaTAM (2024)~\cite{splatam} and MonoGS (2024)~\cite{monogs}, achieving high-fidelity mapping in static environments at substantially lower computational costs. 
Nevertheless, existing 3DGS-SLAM approaches typically assume a static scene, leading to temporally inconsistent updates of Gaussian primitives in dynamic regions, which accumulate over time and contaminate the reconstructed map. 

This behavior results in tracking instability, persistent noise in long-term mapping, and ghosting artifacts, thereby limiting the applicability of 3DGS-SLAM in real-world dynamic scenarios.

In dynamic SLAM systems such as DG-SLAM~\cite{dgslam} and RoDyn-SLAM~\cite{rodyn}, detected dynamic regions are excluded or downweighted during tracking and mapping.
While this strategy reduces the influence of moving objects on camera-pose estimation and map optimization, it does not supply static-background observations for the masked regions. Consequently, backgrounds that remain insufficiently observed, whether under persistent or intermittent occlusion, may be missing or only sparsely represented in the reconstructed map.
When only sparse or inconsistent static observations are available, their reconstructed appearance and geometry may also be weakly constrained, resulting in local blur, ghosting, or unstable Gaussian structures. Therefore, suppression alone removes unreliable dynamic observations but does not provide the missing background information needed for reconstruction.

To address this limitation, our DPGS-SLAM introduces asynchronous diffusion-based background completion together with provenance-aware Gaussian mapping. Here, \textbf{provenance-aware} denotes the explicit labeling of Gaussian primitives as observation-supported or diffusion-generated, with each label determining whether a primitive participates in optimization and how it is updated.

For selected keyframes, a diffusion inpainting model generates plausible background content for regions masked as dynamic, providing an appearance prior where direct observations are unavailable. The completion process runs asynchronously, without blocking camera tracking or observed-map optimization. Metric3D v2 (2024)~\cite{metric3d} predicts metric depth for the completed regions, allowing the generated content to be lifted into 3D and represented by provisional Gaussian primitives.

These generated Gaussian primitives remain fixed and are excluded from camera-pose estimation and observed-map optimization. When the corresponding background becomes visible in subsequent keyframes, a depth-aware projective rule removes eligible generated Gaussian primitives, while observation-supported Gaussian primitives are independently initialized from real RGB observations and predicted depth. In this way, DPGS-SLAM provides provisional representations of insufficiently observed static backgrounds while keeping generated Gaussian primitives separate from camera tracking and observation-supported map optimization.

Our main contributions are as follows:

\begin{itemize}
    \item We design a \textbf{motion-aware dynamic-region identification pipeline} for motion mask generation that integrates periodic optical-flow-based motion discovery with SAM2 (segment anything model 2) (2025)~\cite{sam2} for mask refinement and temporal propagation, supporting dynamic-object discovery and spatiotemporally consistent motion masks.

    \item We introduce an \textbf{asynchronous diffusion-based background completion strategy} for static-background regions with insufficient observation support. A fine-tuned diffusion inpainting model generates plausible background appearance for selected keyframes, while Metric3D v2 (2024)~\cite{metric3d} predicts depth to initialize provisional Gaussian primitives without blocking camera tracking or observed-map optimization.

    \item We propose a \textbf{provenance-aware Gaussian mapping strategy} that explicitly distinguishes observation-supported primitives from diffusion-generated primitives. Generated primitives are excluded from camera-pose estimation and observed-map optimization. As previously occluded backgrounds become visible, eligible generated primitives are selectively removed using a depth-aware projective rule, while observation-supported primitives are independently initialized from real RGB observations and predicted depth.

    \item We assessed DPGS-SLAM on TUM RGB-D~\cite{tum}, BONN RGB-D~\cite{refusion}, and StaDy4D~\cite{stady4d} using monocular RGB input. Our DPGS-SLAM achieved average ATE (absolute trajectory error) RMSE (root mean squared error) values of 1.2\,cm and 2.7\,cm on TUM RGB-D and BONN RGB-D, respectively, demonstrating competitive camera tracking. On the three evaluated StaDy4D sequences, it improved average PSNR (peak signal-to-noise ratio) by 4.60\,dB over WildGS-SLAM (2025)~\cite{wildgs}, while quantitative ablation studies and qualitative comparisons further support the effectiveness of background completion and observation-based replacement.

\end{itemize}

The rest of this paper is organized as follows.
Section~\ref{sec:rel} reviews diffusion-based 3D reconstruction and scene completion, 3D Gaussian splatting SLAM, and dynamic visual SLAM.
Section~\ref{sec:met} presents our proposed method, DPGS-SLAM.
Section~\ref{sec:exp} presents the experimental setup, comparative results, ablation studies, and limitations. Section~\ref{sec:con} concludes this paper.

\begin{figure*}[t]
    \centering
    \includegraphics[
        width=1\textwidth,
        trim={4bp 3bp 122bp 2bp},
        clip
    ]{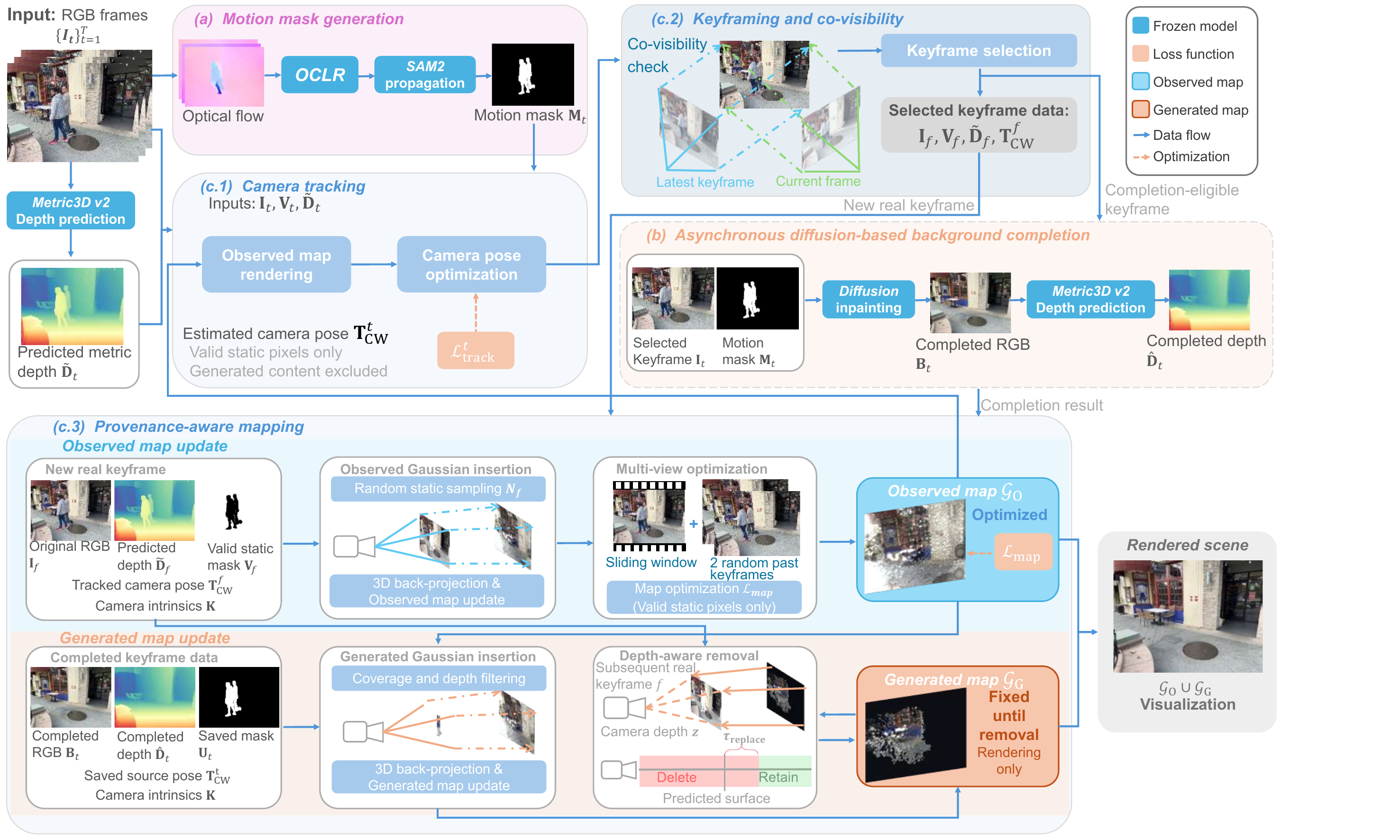}

    \caption{
        \textbf{System overview} -- Our framework consists of (a) motion mask generation (Section~\ref{sec:mask}), (b) asynchronous diffusion-based background completion (Section~\ref{sec:back}), and (c) 3DGS-based SLAM (Section~\ref{sec:slam}), including (c.1) camera tracking, (c.2) keyframing and co-visibility, and (c.3) provenance-aware mapping. 
        Dynamic regions are identified using motion discrepancies revealed by optical flow, with SAM2~\cite{sam2} providing temporally consistent motion masks, while tracking uses observed static regions and background completion provides provisional representations for mapping.
    }
    \label{fig:system_overview}
\end{figure*}

\section{Related Works}
\label{sec:rel}

\subsection{Diffusion-Based 3D Reconstruction and Scene Completion}

Diffusion models learn data distributions through iterative denoising~\cite{diffusion}, while latent diffusion models perform this process in a compressed latent space to enable efficient conditional image generation and inpainting~\cite{latentDiffusion}. Their ability to generate plausible content in unobserved regions has motivated their application to 3D reconstruction. GS-GS (2025)~\cite{gsgs} uses diffusion-generated views to refine Gaussian representations, while Zhong et al. (2025)~\cite{sceneGrounding} introduce scene-grounded video diffusion for sparse-view reconstruction. RI3D (2025)~\cite{ri3d} employs separate diffusion priors for repairing observed regions and inpainting missing content.

Diffusion-based 3D scene completion further addresses missing or occluded regions. NeRFiller (2024)~\cite{nerfiller} uses multi-view diffusion inpainting for NeRF completion, while InFusion (2024)~\cite{infusion} introduces diffusion-based depth completion for initializing missing Gaussian primitives. SplatFill (2026)~\cite{splatfill} combines diffusion inpainting with depth-aware supervision, and Lai et al. (2025)~\cite{lai2025egocentric} employ diffusion-based depth completion for egocentric reconstruction under dynamic occlusions. Although these approaches improve reconstruction completeness, diffusion-generated content may lack direct observation support, requiring careful integration into incremental SLAM to avoid affecting camera tracking and observed-map optimization.

\subsection{3D Gaussian Splatting SLAM Systems}

3DGS~\cite{3dgs} represents scenes as collections of anisotropic Gaussian primitives and enables efficient differentiable rasterization without ray marching.
Owing to its rendering efficiency and reconstruction quality, 3DGS has been widely adopted in dense SLAM systems, including SplaTAM (2024)~\cite{splatam}, MonoGS (2024)~\cite{monogs}, and Photo-SLAM (2024)~\cite{photo}, achieving accurate tracking and high-quality mapping in static environments. However, these systems are primarily developed and evaluated under static-scene assumptions.

To handle dynamic environments, recent works extend 3DGS-SLAM to dynamic scenarios, including DG-SLAM (2024)~\cite{dgslam} and GassiDy (2025)~\cite{GassiDy}. These approaches identify and suppress dynamic observations through motion masks, rendering residuals, or adaptive Gaussian primitive management.
WildGS-SLAM (2025)~\cite{wildgs} further incorporates learned per-pixel uncertainty into tracking and Gaussian map optimization to reduce the influence of dynamic observations. DRG-SLAM (2026)~\cite{drgslam} combines geometric motion verification and temporal consistency to exclude dynamic observations from RGB-D tracking and Gaussian mapping.
While these strategies reduce contamination from moving objects, dynamic suppression alone does not provide observation support for persistently occluded static backgrounds. As a result, such regions may remain missing, sparsely represented, or weakly constrained in the reconstructed Gaussian map.


\subsection{Dynamic Visual SLAM Systems}

Many classical visual SLAM systems, such as DSO~\cite{dso} and ORB-SLAM3 (2021)~\cite{orbslam3}, are developed under the assumption of static scenes.
The presence of dynamic objects contradicts this assumption by disrupting feature correspondences and photometric consistency, often resulting in tracking drift and reconstruction failure.
To address this limitation, a common strategy is to identify and exclude dynamic regions, reconstructing only the static components of the environment.

Dynamic region detection is commonly based on geometric consistency checks, such as StaticFusion (2018)~\cite{staticfusion} and ReFusion~\cite{refusion}, or optical-flow-based motion estimation, such as Motion removal (2018)~\cite{motionremoval} and FlowFusion (2020)~\cite{flowfusion}.
However, these approaches can produce noisy and temporally unstable masks under occlusions or inaccurate geometry and motion estimates.
Semantic-based methods further exploit object detection or semantic segmentation priors like Mask-SLAM (2018)~\cite{maskslam}, but semantic category information alone cannot reliably distinguish between moving and stationary instances.
Hybrid approaches combining geometric, motion, and semantic cues, such as DynaSLAM (2018)~\cite{dynaslam} and SG-SLAM (2022)~\cite{sgslam}, improve dynamic-region identification, although consistent separation of dynamic and static regions remains challenging under occlusions and intermittent observations.
More recently, DROID-W (2026)~\cite{droidw} introduces uncertainty-aware bundle adjustment to jointly refine camera poses, depth, and dynamic uncertainty.

In many dynamic SLAM systems~\cite{dynaslam, rodyn, dgslam}, dynamic observations are excluded or suppressed during tracking and mapping.
While hard exclusion prevents detected dynamic pixels from directly contributing to pose estimation and map optimization, masking alone does not provide static-background information for regions that remain unobserved.

Beyond hard exclusion, Zhou et al. (2025)~\cite{zhou2025dynamic} combine dynamic pixel-set masking with residual losses for soft suppression. Dy3DGS-SLAM (2025)~\cite{li2025dy3dgs} addresses artifacts associated with pruning dynamic Gaussian primitives. These strategies reduce dynamic interference but do not supply missing static-background observations; soft suppression may also retain residual dynamic contamination. 

DPGS-SLAM therefore combines optical-flow-based motion discovery with temporally propagated segmentation and performs asynchronous diffusion-based background completion to provisionally represent backgrounds with insufficient real observation support. Provenance-aware mapping isolates generated Gaussian primitives from camera-pose estimation and observed-map optimization; when real static observations become available, eligible generated primitives are removed and observation-supported primitives are independently initialized.

\begin{figure}[t]
    \centering
    \includegraphics[
        width=0.98\linewidth,
        trim={0bp 26bp 643bp 0bp},
        clip
    ]{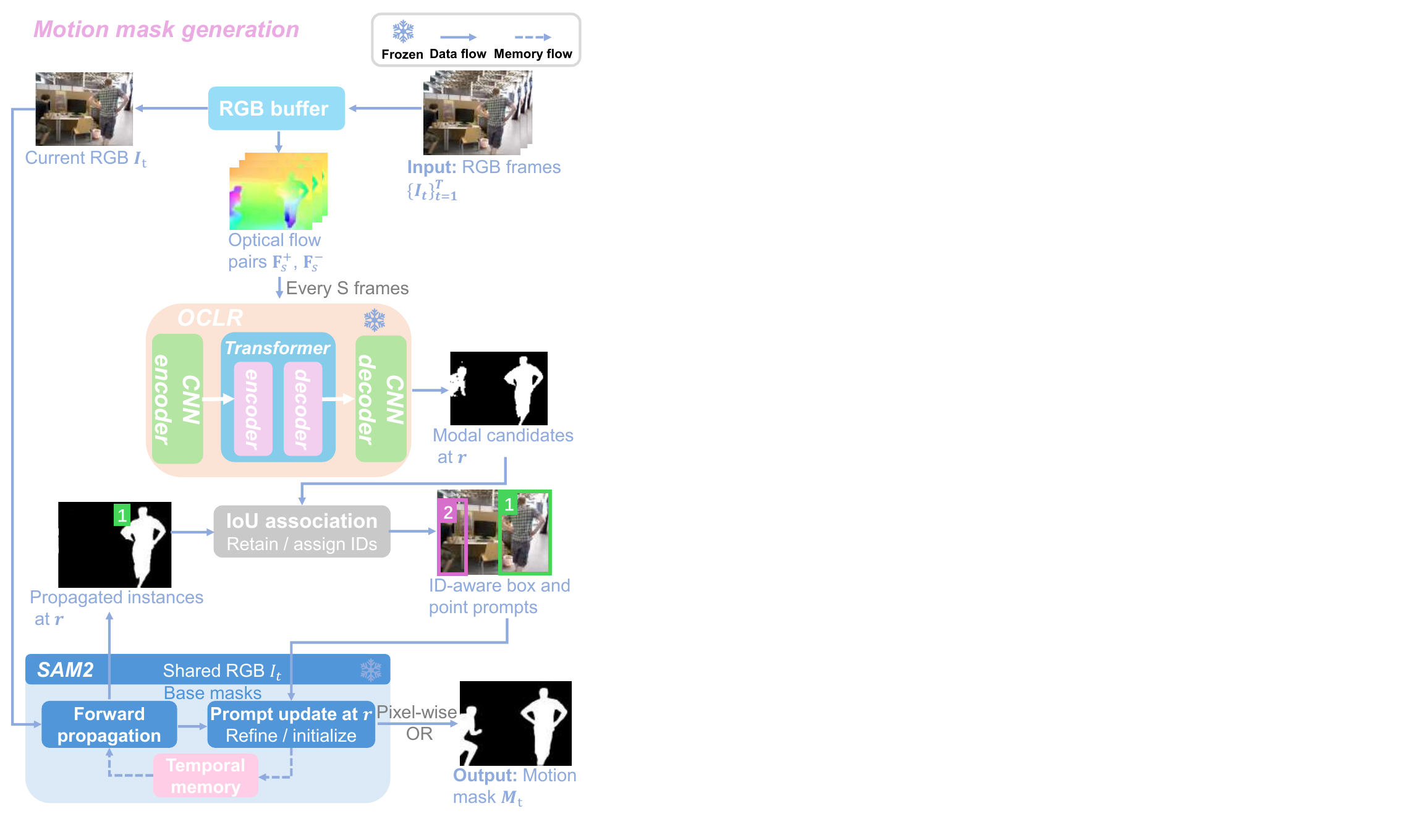}
    \vspace{-5pt}
    \caption{\textbf{Motion mask generation}
        (Section~\ref{sec:mask}): 
        OCLR~\cite{oclr} periodically identifies motion candidates from local optical-flow windows. Association with SAM2~\cite{sam2} predictions generates ID-aware box and point prompts for refining existing masks and initializing new objects. SAM2 propagates registered objects forward to produce temporally consistent dynamic-region masks.}
    \label{fig:motion_mask_generation}
\end{figure}


\section{Methodology}\label{sec:met}

The overall pipeline is illustrated in Fig.~\ref{fig:system_overview}, which
consists of (a) motion mask generation (Section~\ref{sec:mask}), (b) asynchronous diffusion-based background completion (Section~\ref{sec:back}), and (c) 3DGS-based SLAM (Section~\ref{sec:slam}).


\subsection{Motion Mask Generation}
\label{sec:mask}

The process of motion mask generation, roughly shown in Fig.~\ref{fig:system_overview}~(a), is further presented in greater detail in Fig.~\ref{fig:motion_mask_generation}. 
Given RGB frames $\{\mathbf{I}_t\}_{t=1}^{T}$, where $\mathbf{I}_t\in\mathbb{R}^{H\times W\times 3}$, we estimate binary masks $\mathbf{M}_t\in\{0,1\}^{H\times W}$ identifying dynamic foreground regions to exclude from static-background reconstruction. Motion discovery is performed every $S$ frames using only RGB input, independently of SLAM pose estimates and keyframe selection.

At each detection frame $r$ satisfying $\Delta+1\leq r\leq T-1$, a pretrained OCLR (object-centric layered representation) (2022)~\cite{oclr} model processes $\Delta$ consecutive pairs of optical flow fields:
\begin{equation}
\begin{aligned}
    \mathbf{F}_s^{+}
    &= \mathcal{F}(\mathbf{I}_s,\mathbf{I}_{s+1}),\\
    \mathbf{F}_s^{-}
    &= \mathcal{F}(\mathbf{I}_s,\mathbf{I}_{s-1}),
\end{aligned}
\label{eq:motion_flow}
\end{equation}
where $s\in\{r-\Delta+1,\ldots,r\}$ and $\mathcal{F}$ is a pretrained optical flow estimator. Each pair shares the source-frame grid, and the forward and backward two-channel flow fields are concatenated into a four-channel input per frame, matching the input configuration of the pretrained OCLR network. 
Amodal predictions (of partially hidden or occluded objects) by OCLR are composed using its estimated layer ordering; only the resulting modal masks at frame $r$ are used as object candidates.

Candidates are associated with propagated masks by SAM2~\cite{sam2} at frame $r$ by greedy one-to-one matching in descending order of intersection over union (IoU), rejecting pairs below an IoU threshold. Matched objects retain their IDs, while unmatched candidates receive new IDs.
Candidate bounding boxes and interior positive points prompt SAM2 to refine or initialize the corresponding masks. The refined masks are used for the current frame and subsequent forward propagation. Once registered as dynamic, objects remain masked during temporary stops while segmented as visible. Let $\mathcal{O}_t$ denote the object IDs registered up to frame $t$ and $\mathbf{M}_t^{(i)}$ its estimated visible mask, set to zero when no visible region is predicted. The combined mask is:
\begin{equation}
    \mathbf{M}_t
    = \bigvee_{i\in\mathcal{O}_t}\mathbf{M}_t^{(i)},
    \label{eq:motion_mask_union}
\end{equation}
where $\bigvee$ denotes pixel-wise logical OR; an empty union yields an all-zero mask.

A complete flow window requires $\Delta+2$ RGB frames; detection at $r$ therefore waits for $\mathbf{I}_{r+1}$. SLAM starts at the first detection frame $t_{\mathrm{init}}$, with preceding frames used only for motion initialization. Image--mask pairs are submitted chronologically after their masks are finalized, and submitted masks are not revised. After the last detection, SAM2 propagates already registered objects over the remaining frames. The static observation mask is:
\begin{equation}
    \mathbf{V}_t
    = \mathbf{V}^{\mathrm{rgb}}_t
      \odot\mathbf{V}^{\mathrm{depth}}_t
      \odot(1-\mathbf{M}_t),
    \label{eq:staticmask}
\end{equation}
where $\mathbf{V}^{\mathrm{rgb}}_t$ and $\mathbf{V}^{\mathrm{depth}}_t$ indicate valid RGB pixels and valid monocular depth predictions, respectively, and $\odot$ denotes element-wise multiplication.

\begin{figure}[t]
  \centering
    \includegraphics[
        width=0.98\linewidth,
        trim={5bp 100bp 650bp 0bp},
        clip
    ]{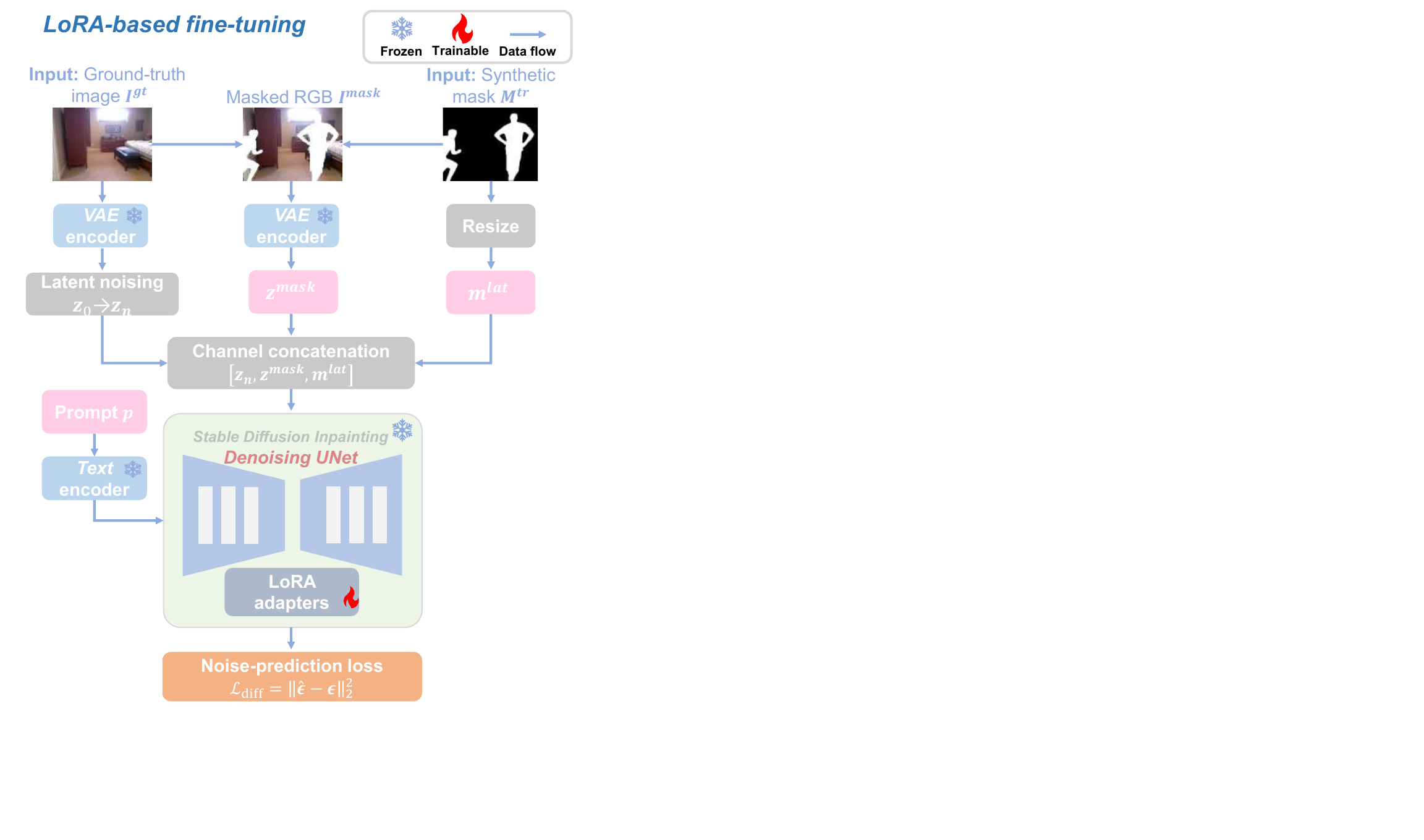}
    \vspace{-5pt}

    \caption{
        \textbf{LoRA-based fine-tuning} (Section~\ref{sec:back:training}).
        Static background images and synthetic occlusion masks
        provide training pairs for the Stable Diffusion inpainting model.
        Only the LoRA~\cite{lora} adapters are optimized using the
        noise-prediction loss, while the base U-Net, VAE,
        and text encoder remain frozen.
    }
    \label{fig:completion_training}
\end{figure}


\subsection{Asynchronous Diffusion-Based Background Completion} \label{sec:back}

The process of asynchronous diffusion-based background completion roughly shown in Fig.~\ref{fig:system_overview}~(b) is explained in detail. 

\subsubsection{Model and LoRA-Based Fine-Tuning} \label{sec:back:training}
We fine-tuned a latent diffusion inpainting model initialized from Stable Diffusion Inpainting (2022)~\cite{latentDiffusion} to reconstruct static-background content in motion-masked regions. The variational autoencoder (VAE) encoder $\mathcal{E}_{\mathrm{VAE}}$, VAE decoder $\mathcal{D}_{\mathrm{VAE}}$, text encoder $\mathcal{T}$, and the pretrained denoising U-Net parameters $\theta_0$ were frozen, while only the Low-Rank Adaptation (LoRA) (2022)~\cite{lora} parameters $\phi$ inserted into the U-Net were optimized. During inference, diffusion is invoked only for selected keyframes and is executed asynchronously in an independent worker, so camera tracking does not synchronously wait for completion.
In the following, two-dimensional masks are broadcast across RGB channels, and $1-\mathbf{M}$ denotes the element-wise mask complement. Norms of image and latent arrays are evaluated after vectorization.

Training samples were constructed from static background images using synthetic occlusion masks. The LoRA fine-tuning pipeline is illustrated in Fig.~\ref{fig:completion_training}. Given a ground-truth (GT) static image $\mathbf{I}^{\mathrm{gt}}\in\mathbb{R}^{H\times W\times 3}$ and a sampled binary training mask $\mathbf{M}^{\mathrm{tr}}\in\{0,1\}^{H\times W}$, the masked image is:

\begin{equation}
    \mathbf{I}^{\mathrm{mask}}
    =
    (1-\mathbf{M}^{\mathrm{tr}})
    \odot
    \mathbf{I}^{\mathrm{gt}}.
    \label{eq:diff-masked-input}
\end{equation}

The ground-truth and masked images are encoded into latent space as:
\begin{equation}
    \mathbf{z}_0
    =
    \mathcal{E}_{\mathrm{VAE}}
    \left(\mathbf{I}^{\mathrm{gt}}\right),
    \qquad
    \mathbf{z}^{\mathrm{mask}}
    =
    \mathcal{E}_{\mathrm{VAE}}
    \left(\mathbf{I}^{\mathrm{mask}}\right).
    \label{eq:diff-latents}
\end{equation}
Both latent arrays have dimensions $H_z\times W_z\times C_z$, where $H_z$, $W_z$, and $C_z$ denote the latent height, width, and number of channels, respectively.

The binary training mask is resized to the latent resolution as:
\begin{equation}
    \mathbf{m}^{\mathrm{lat}}
    =
    \mathcal{S}\left(\mathbf{M}^{\mathrm{tr}}\right),
    \label{eq:latent-mask}
\end{equation}
where $\mathcal{S}(\cdot)$ denotes spatial resampling and $\mathbf{m}^{\mathrm{lat}}\in[0,1]^{H_z\times W_z\times 1}$.

We use an $\epsilon$-prediction diffusion formulation. At diffusion timestep $n$, Gaussian noise is added to the ground-truth latent as:
\begin{equation}
    \mathbf{z}_n
    =
    \sqrt{\bar{\alpha}_n}\mathbf{z}_0
    +
    \sqrt{1-\bar{\alpha}_n}\boldsymbol{\epsilon}.
    \label{eq:forward-diffusion}
\end{equation}
Here, $\bar{\alpha}_n=\prod_{j=1}^{n}\alpha_j$ is the cumulative signal-retention coefficient determined by the diffusion noise schedule, and $\alpha_j$ denotes the signal-retention coefficient at diffusion timestep $j$. The noise $\boldsymbol{\epsilon}$ has the same shape as $\mathbf{z}_0$, with entries independently sampled from $\mathcal{N}(0,1)$.

Let $\mathbf{c}=\mathcal{T}(p)$ denote the text-conditioning embedding of prompt $p$. Following the inpainting formulation, the denoising U-Net receives the noisy latent together with the masked-image latent and the latent-resolution mask:
\begin{equation}
    \hat{\boldsymbol{\epsilon}}
    =
    \epsilon_{\theta_0,\phi}
    \left(
        [\mathbf{z}_n,\mathbf{z}^{\mathrm{mask}},\mathbf{m}^{\mathrm{lat}}],
        n,
        \mathbf{c}
    \right),
    \label{eq:inpainting-unet}
\end{equation}
where $[\cdot]$ denotes channel-wise concatenation. The LoRA parameters are optimized using the standard diffusion noise-prediction objective:
\begin{equation}
    \mathcal{L}_{\mathrm{diff}}
    =
    \left\|
        \hat{\boldsymbol{\epsilon}}
        -
        \boldsymbol{\epsilon}
    \right\|_2^2.
    \label{eq:diffusion}
\end{equation}

\begin{figure}[t]
  \centering
    \includegraphics[
        width=0.98\linewidth,
        trim={0bp 118bp 643bp 0bp},
        clip
    ]{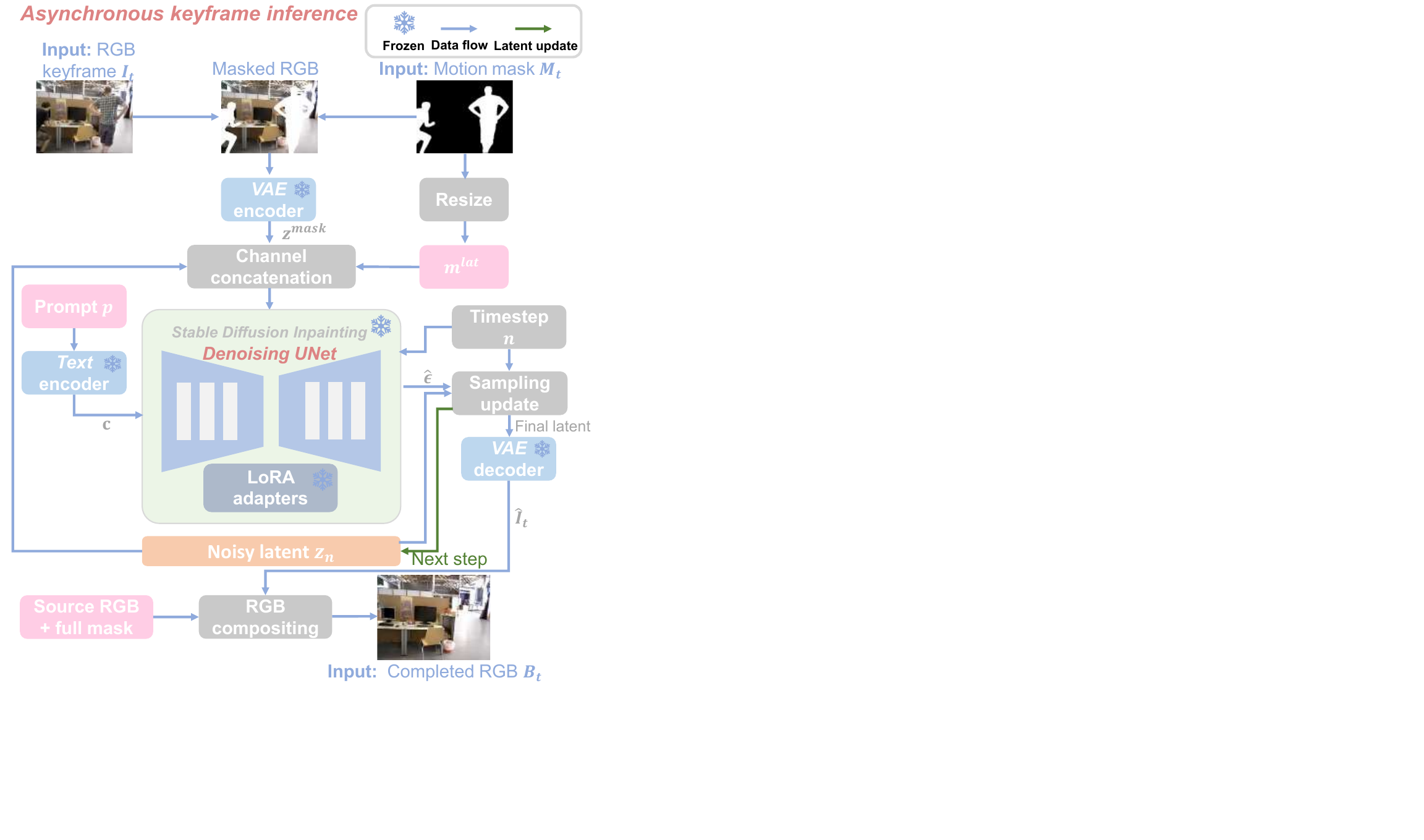}
    \vspace{-5pt}

    \caption{
        \textbf{Asynchronous keyframe inference} (Section~\ref{sec:back:inference}).
        Eligible keyframes undergo asynchronous background completion using the LoRA-adapted inpainting model with all model parameters fixed. After VAE decoding, compositing retains generated pixels within the full motion mask and preserves original pixels elsewhere. Tracking and observed-map optimization proceed without waiting for completion, and completed frames are excluded from camera-pose optimization.
    }
    \label{fig:diff_inference}
\end{figure}

\subsubsection{Asynchronous Keyframe Inference} \label{sec:back:inference}
Background completion is invoked only for selected keyframes that contain sufficiently large dynamic regions with insufficient coverage from the current Gaussian map. The asynchronous background-completion pipeline is illustrated in Fig.~\ref{fig:diff_inference}. Let $\mathbf{C}_t\in\{0,1\}^{H\times W}$ denote its rendered coverage at the frame pose. For single-channel maps, $A(u,v)$ denotes the value of array $\mathbf{A}$ at horizontal coordinate $u$ and vertical coordinate $v$ in the pixel domain $\Omega=\{1,\ldots,W\}\times\{1,\ldots,H\}$. Coverage is defined by:
\begin{equation}
    C_t(u,v)
    =\mathbb{I}\!\left[
    A_t^{\mathrm{all}}(u,v)\geq\tau_{\mathrm{cov}}
    \right],
    \label{eq:map-coverage}
\end{equation}
where $\mathbf{A}_t^{\mathrm{all}}\in[0,1]^{H\times W}$ is the accumulated-opacity image rendered from the combined observed and generated Gaussian map $\mathcal{G}_{\mathrm O}\cup\mathcal{G}_{\mathrm G}$, $\tau_{\mathrm{cov}}\in(0,1]$ is the coverage threshold, and $\mathbb{I}[\cdot]$ is the indicator function. We define the masked region with insufficient map coverage as:
\begin{equation}
    \mathbf{U}_t
    =
    \mathbf{M}_t
    \odot
    \left(
        \mathbf{1}_{H\times W}-\mathbf{C}_t
    \right),
    \label{eq:unknown-dynamic-region}
\end{equation}
where $\mathbf{1}_{H\times W}$ is the all-one array and $U_t(u,v)=1$ indicates a pixel with insufficient map coverage. Its normalized area is:
\begin{equation}
    a_t
    =
    \frac{1}{HW}
    \sum_{v=1}^{H}\sum_{u=1}^{W}U_t(u,v).
    \label{eq:completion-area}
\end{equation}
A diffusion task is scheduled only if:
\begin{equation}
    a_t\geq\tau_{\mathrm{comp}},
    \label{eq:completion-trigger}
\end{equation}
where $\tau_{\mathrm{comp}}$ is the minimum unknown-area threshold.

For each selected keyframe, the full-resolution masked image $(1-\mathbf{M}_t)\odot\mathbf{I}_t$ and its motion mask $\mathbf{M}_t$ are passed to the diffusion inpainting model, producing an inpainted image $\hat{\mathbf{I}}_t$. Only the generated pixels within the masked region are retained:
\begin{equation}
    \mathbf{B}_t
    =
    \mathbf{M}_t\odot\hat{\mathbf{I}}_t
    +
    (1-\mathbf{M}_t)\odot\mathbf{I}_t.
    \label{eq:blend}
\end{equation}
Thus, all unmasked pixels remain unchanged. Completion is executed asynchronously, and the tracking and observed-map optimization loops do not wait for its result. The background-completed frame $\mathbf{B}_t$ is never used for camera pose estimation.

\subsection{Provenance-Aware 3DGS-SLAM Pipeline}
\label{sec:slam}

\subsubsection{3D Gaussian-Based Map Representation}
We represent the scene using the 3DGS formulation, in which the environment is modeled as a set of anisotropic 3D Gaussian primitives:
\begin{equation}
    \mathcal{G}
    =
    \{G_i=(\boldsymbol{\mu}_i,\mathbf{\Sigma}_i,o_i,\mathbf{h}_i,q_i)
    \mid i=1,\dots,N\}.
\end{equation}
Here, $N$ is the current number of primitives. Each Gaussian primitive is parameterized by its center $\boldsymbol{\mu}_i\in\mathbb{R}^3$, covariance $\mathbf{\Sigma}_i\in\mathbb{R}^{3\times3}$, scalar opacity $o_i\in(0,1)$, and spherical harmonic coefficients $\mathbf{h}_i\in\mathbb{R}^{3\times(L+1)^2}$ representing view-dependent appearance, where $L$ denotes the maximum order of the spherical harmonics expansion~\cite{3dgs}. The provenance label $q_i\in\{\mathrm{OBSERVED},\mathrm{GENERATED}\}$ distinguishes primitives initialized from real static RGB pixels and diffusion-generated RGB pixels. Accordingly, we denote the observed and generated Gaussian subsets by $\mathcal{G}_{\mathrm{O}}$ and $\mathcal{G}_{\mathrm{G}}$, respectively, with $\mathcal{G}=\mathcal{G}_{\mathrm{O}}\cup\mathcal{G}_{\mathrm{G}}$.

Given a camera pose $\mathbf{T}_{\mathrm{CW}}^t\in\mathrm{SE}(3)$ (rotation + translation), world-frame primitives are transformed into the camera frame and rendered using the standard differentiable rasterization and alpha-compositing procedure of 3DGS~\cite{3dgs}. The subscript $\mathrm{CW}$ denotes the world-to-camera transformation, and $\mathrm{WC}$ denotes its inverse.

\begin{figure}[t]
    \centering
    \includegraphics[
        width=\linewidth,
        trim={12bp 536bp 881bp 2bp},
        clip
    ]{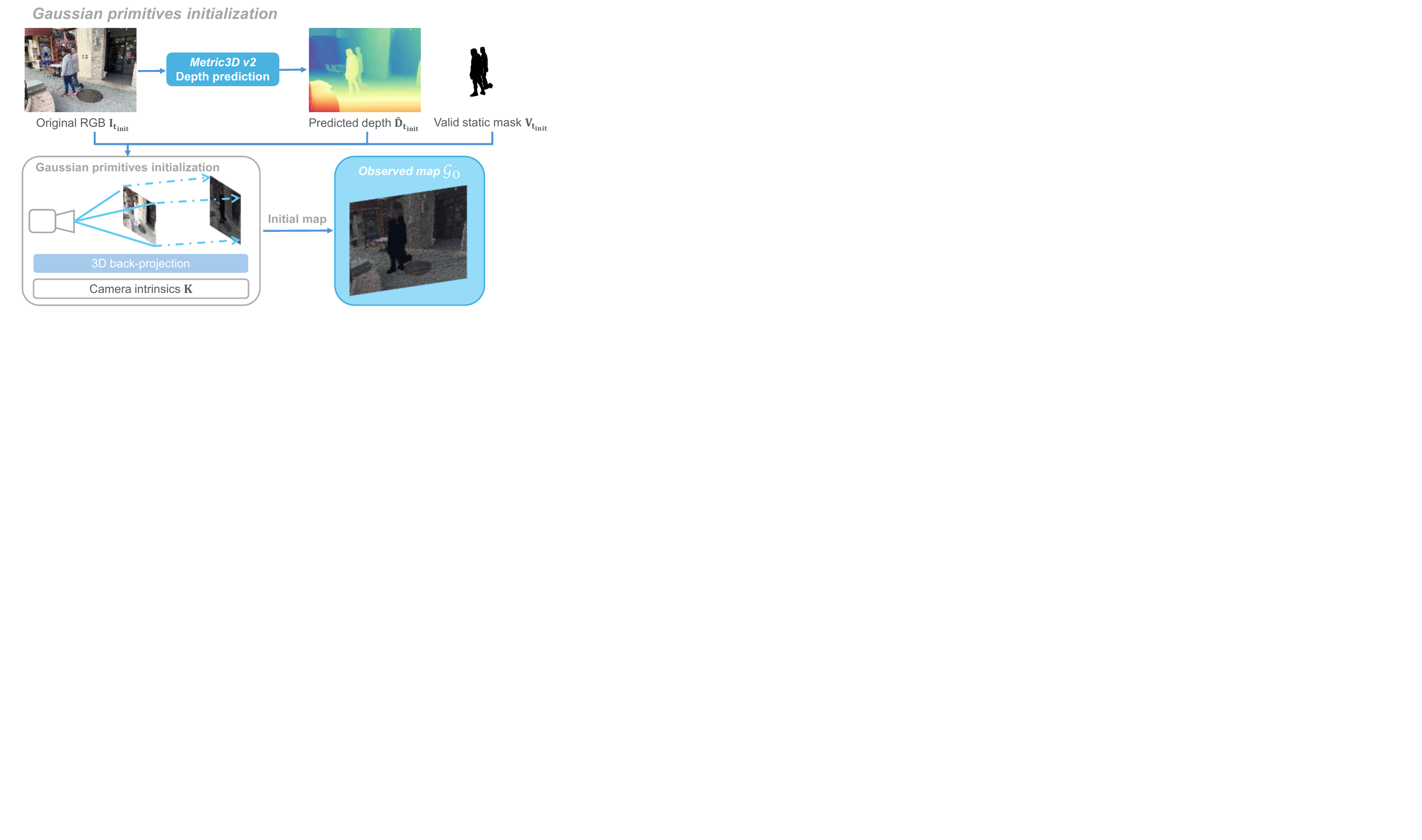}

    \caption{
        \textbf{Gaussian primitives initialization}
        (Section~\ref{sec:init}).
        At the initial SLAM frame, Metric3D v2 predicts depth
        from the original RGB image.
        Valid static pixels are back-projected using the predicted
        depth and camera intrinsics to initialize the observed
        Gaussian map $\mathcal{G}_{\mathrm{O}}$.
    }
    \label{fig:gaussian_initialization}
\end{figure}

\subsubsection{Gaussian Primitives Initialization}
\label{sec:init}
For the initial SLAM frame at time $t_{\mathrm{init}}$, inspired by WildGS-SLAM~\cite{wildgs}, we initialize the observation-supported Gaussian map from valid static pixels selected by $\mathbf{V}_{t_{\mathrm{init}}}$. Metric depth $\tilde{\mathbf{D}}_{t_{\mathrm{init}}}\in\mathbb{R}^{H\times W}$ is predicted from the original frame $\mathbf{I}_{t_{\mathrm{init}}}$ using Metric3D v2~\cite{metric3d}, and the valid pixels are back-projected using the camera intrinsics to initialize $\mathcal{G}_{\mathrm{O}}$. The resulting primitives are assigned $q_i=\mathrm{OBSERVED}$ and provide the starting point cloud for subsequent tracking and mapping (see Fig.~\ref{fig:gaussian_initialization}). All depth maps represent camera-frame $z$-coordinates, with finite positive predictions selected by the depth-validity mask.

\subsubsection{Camera Tracking}
\label{sec:tracking}
Given a new frame at time $t$, we optimize only the camera pose $\mathbf{T}_{\mathrm{CW}}^t$, while keeping all Gaussian parameters fixed (see Fig.~\ref{fig:system_overview}~(c.1)). 
Only observation-supported Gaussian primitives and valid static pixels are used for pose estimation:
\begin{align}
    \mathcal{L}_{\mathrm{pho}}^t
    &=
    \left\|
        \mathbf{V}_t\odot
        \left[
            \mathcal{R}_{\mathrm{rgb}}
            \left(
                \mathcal{G}_{\mathrm{O}},
                \mathbf{T}_{\mathrm{CW}}^t
            \right)
            -\mathbf{I}_t
        \right]
    \right\|_1,
    \label{eq:pho}\\
    \mathcal{L}_{\mathrm{geo}}^t
    &=
    \left\|
        \mathbf{V}_t\odot
        \left[
            \mathcal{R}_{\mathrm{depth}}
            \left(
                \mathcal{G}_{\mathrm{O}},
                \mathbf{T}_{\mathrm{CW}}^t
            \right)
            -\tilde{\mathbf{D}}_t
        \right]
    \right\|_1.
    \label{eq:geo}
\end{align}
Here, $\mathcal{R}_{\mathrm{rgb}}$ and $\mathcal{R}_{\mathrm{depth}}$ produce color and depth arrays of dimensions $H\times W\times3$ and $H\times W$, respectively, and $\tilde{\mathbf{D}}_t$ is predicted from the original frame $\mathbf{I}_t$ by Metric3D v2~\cite{metric3d}. 
The final tracking objective is:
\begin{equation}
    \mathcal{L}_{\mathrm{track}}^t
    =
    \lambda_{\mathrm{pho}}\mathcal{L}_{\mathrm{pho}}^t
    +
    \lambda_{\mathrm{geo}}\mathcal{L}_{\mathrm{geo}}^t,
    \label{eq:lossTrack}
\end{equation}
where $\lambda_{\mathrm{pho}}$ and $\lambda_{\mathrm{geo}}$ balance the photometric and geometric residuals. Camera pose optimization is performed using gradient descent for a fixed number of iterations $N_{\mathrm{track}}$ per frame. Generated Gaussian primitives and dynamic pixels do not contribute gradients to camera pose optimization.

\subsubsection{Keyframing and Co-Visibility}
\label{sec:keyframe_co}
To avoid optimizing all frames, we maintain a fixed-size sliding window $\mathcal{W}_{\mathrm c}$ of keyframe indices selected based on a co-visibility criterion (see Fig.~\ref{fig:system_overview}~(c.2)). 
For each incoming frame, we estimate co-visibility by measuring the intersection over union (IoU) of the sets of visible observation-supported Gaussian primitives between the current frame and the latest keyframe. If this score falls below a threshold, or if the camera has moved significantly relative to the median rendered depth of the observed map in the latest keyframe, the current frame is selected as a keyframe.

Real static pixels from a newly selected keyframe are immediately sent to the mapping backend. Keyframes satisfying the completion criterion in Section~\ref{sec:back} are additionally submitted to the asynchronous diffusion worker.

\subsubsection{Provenance-Aware Mapping}
\label{sec:mapping}
During mapping, camera poses estimated during tracking are fixed, and $\mathcal{G}_{\mathrm{O}}$ is optimized using multi-view static observations (see Fig.~\ref{fig:system_overview}~(c.3)). 
Following the predicted-depth-guided Gaussian initialization strategy of WildGS-SLAM~\cite{wildgs}, valid static pixels from each new keyframe $f$ are randomly subsampled:
\begin{equation}
    \mathbf{N}_f
    =\operatorname{Sample}_{\rho}(\mathbf{V}_f),
    \label{eq:observed-insertion}
\end{equation}
where $\operatorname{Sample}_{\rho}$ selects a fraction $\rho$ of valid static pixels. Selected pixels are back-projected using predicted depth $\tilde{\mathbf{D}}_f$ and the estimated camera pose to initialize new Gaussian primitives with $q_i=\mathrm{OBSERVED}$, with appearance initialized from the original RGB image $\mathbf{I}_f$. The same procedure applies to both ordinary static regions and newly revealed backgrounds.

We use the current sliding window $\mathcal{W}_{\mathrm c}$ together with two randomly sampled past keyframe indices $\mathcal{W}_{\mathrm r}$ to reduce forgetting of the global map, and define $\mathcal{W}=\mathcal{W}_{\mathrm c}\cup\mathcal{W}_{\mathrm r}$. Reusing the photometric and geometric losses in~\eqref{eq:pho} and~\eqref{eq:geo}, the mapping objective is:
\begin{equation}
    \begin{aligned}
        \mathcal{L}_{\mathrm{map}}
        ={}&
        \sum_{f\in\mathcal{W}}
        \left(
            \lambda_{\mathrm{pho}}\mathcal{L}_{\mathrm{pho}}^f
            +
            \lambda_{\mathrm{geo}}\mathcal{L}_{\mathrm{geo}}^f
        \right)\\
        &+
        \lambda_{\mathrm{iso}}
        \sum_{G_i\in\mathcal{G}_{\mathrm{O}}}
        \left\|
            \mathbf{s}_i-\bar{s}_i\mathbf{1}_3
        \right\|_1,
    \end{aligned}
    \label{eq:lossMap}
\end{equation}
where $\mathbf{s}_i=[s_{i,x},s_{i,y},s_{i,z}]^\top\in\mathbb{R}^3$ denotes the anisotropic scaling vector, $\bar{s}_i=\tfrac{1}{3}(s_{i,x}+s_{i,y}+s_{i,z})$ is its mean, and $\lambda_{\mathrm{iso}}$ controls the isotropic regularization strength, following MonoGS (2024)~\cite{monogs}. The same residuals depend on both the map and camera poses: tracking optimizes the poses, whereas mapping optimizes the observed Gaussian parameters. Unlike MonoGS, we retain a geometric loss under monocular RGB input using depth predicted by Metric3D v2 (2024)~\cite{metric3d}.

When an asynchronous completion becomes available, Metric3D v2~\cite{metric3d} predicts depth from the completed frame:
\begin{equation}
    \hat{\mathbf{D}}_t
    =
    \mathrm{Metric3D}(\mathbf{B}_t).
    \label{eq:completed-depth}
\end{equation}
Here, $t$ denotes the source keyframe index, irrespective of when the result becomes available. The source mask $\mathbf{U}_t$ and pose are saved when the task is submitted. At integration, coverage is recomputed from the current map at the saved pose; only pixels in $\mathbf{U}_t$ that remain uncovered according to the recomputed coverage mask and have finite positive completed-depth predictions are used to initialize generated geometry.

For a pixel $(u,v)$ in this region, let $\bar{\boldsymbol{x}}=[u,v,1]^\top$ be its homogeneous image coordinate. The camera- and world-frame points are:
\begin{equation}
    \begin{aligned}
        \hat{\boldsymbol{p}}_{\mathrm c,t}(u,v)
        &=
        \hat D_t(u,v)\,
        \mathbf{K}^{-1}\bar{\boldsymbol{x}},\\
        \hat{\boldsymbol{p}}_{\mathrm w,t}(u,v)
        &=
        \mathbf{R}_{\mathrm{WC}}^{t}
        \hat{\boldsymbol{p}}_{\mathrm c,t}(u,v)
        +\boldsymbol{t}_{\mathrm{WC}}^{t}.
    \end{aligned}
    \label{eq:generated-backprojection}
\end{equation}
Here, $\mathbf{K}\in\mathbb{R}^{3\times3}$ is the intrinsic matrix for the image resolution used, and $\mathbf{R}_{\mathrm{WC}}^{t}\in\mathrm{SO}(3)$ (rotation) and $\boldsymbol{t}_{\mathrm{WC}}^{t}\in\mathbb{R}^3$ are the rotation and translation of $\mathbf{T}_{\mathrm{WC}}^{t}=(\mathbf{T}_{\mathrm{CW}}^{t})^{-1}$. The points initialize provisional Gaussian primitives in $\mathcal{G}_{\mathrm{G}}$, with appearance taken from $\mathbf{B}_t$ and provenance $q_i=\mathrm{GENERATED}$. These primitives remain fixed until they are removed when valid static observations become available and are excluded from camera pose optimization and the observed-map loss.

Generated primitives are removed using a depth-aware projective rule when subsequent keyframes provide valid static observations. For a generated primitive $G_i$ and a subsequent keyframe $f$, its camera-frame center and depth are:
\begin{equation}
    \boldsymbol{p}_{i,f}
    =\mathbf{R}_{\mathrm{CW}}^f\boldsymbol{\mu}_i
    +\boldsymbol{t}_{\mathrm{CW}}^f,
    \qquad z_{i,f}=[\boldsymbol{p}_{i,f}]_3,
    \label{eq:replacement-projection}
\end{equation}
where $\mathbf{R}_{\mathrm{CW}}^f$ and $\boldsymbol{t}_{\mathrm{CW}}^f$ are the rotation and translation of $\mathbf{T}_{\mathrm{CW}}^f$. For $z_{i,f}>0$, let $\boldsymbol{u}_{i,f}=\operatorname{round}(\pi(\mathbf{K}\boldsymbol{p}_{i,f}))$ be its nearest projected pixel, where $\pi([x,y,z]^\top)=[x/z,y/z]^\top$ and rounding is component-wise. Let $\Omega$ denote the valid image pixel domain. The generated primitives to be deleted are:
\begin{equation}
    \mathcal{G}_{\mathrm G}^{\mathrm{del}}(f)
    =\left\{G_i\in\mathcal{G}_{\mathrm G}\;\middle|\;
    \begin{aligned}
        &z_{i,f}>0,\quad \boldsymbol{u}_{i,f}\in\Omega,\\
        &V_f(\boldsymbol{u}_{i,f})=1,\\
        &z_{i,f}\leq\tilde D_f(\boldsymbol{u}_{i,f})
                       +\tau_{\mathrm{replace}}
    \end{aligned}\right\},
    \label{eq:generated-replacement}
\end{equation}
where $\tau_{\mathrm{replace}}\geq0$ is a depth tolerance in meters. This criterion removes generated primitives whose projected centers lie at or in front of the predicted static surface, or within $\tau_{\mathrm{replace}}$ behind it. Primitives farther behind the surface are retained, as their geometry remains unverified from the current view. The criterion uses projected Gaussian centers as an approximation of primitive-level visibility.
Newly revealed static pixels are reconstructed through~\eqref{eq:observed-insertion}, while existing observed primitives are refined by~\eqref{eq:lossMap}. New observed primitives use only real-frame RGB and predicted depth and do not inherit parameters from deleted generated primitives.

Asynchronous background completion provides provisional representations of occluded static backgrounds, with eligible generated primitives selectively removed as valid static observations become available. The combined map $\mathcal{G}_{\mathrm{O}}\cup\mathcal{G}_{\mathrm{G}}$ is used for scene visualization, while only the observation-supported map $\mathcal{G}_{\mathrm{O}}$ is used for camera tracking and observed-map optimization.



\newsavebox{\trackingtabbox}
\begin{table*}[t]
\centering
\scriptsize
\caption{Camera tracking accuracy measured by ATE RMSE ($\downarrow$ [cm]) on BONN RGB-D and TUM RGB-D.
}
\label{tab:tracking_results}

\begin{lrbox}{\trackingtabbox}
\begin{tabular}{c l c | cccc c | cccc c}
\toprule
\textbf{Paradigm} & \textbf{Methods} & \textbf{Type}
& \multicolumn{5}{c|}{\textbf{BONN RGB-D}~\cite{refusion}}
& \multicolumn{5}{c}{\textbf{TUM RGB-D}~\cite{tum}} \\
\cmidrule(lr){4-8} \cmidrule(lr){9-13}
& & & ps.tr & ps.tr2 & balloon & mv.box2 & \textbf{Avg.}
& wk.xyz & wk.hf & wk.st & st.hf & \textbf{Avg.} \\
\midrule


\multirow{3}{*}{\parbox{1.6cm}{\centering Traditional\\ SLAM methods}}
& DynaSLAM (2018)~\cite{dynaslam} & S
& 6.1 & 7.8 & 3.0 & 3.9 & 5.2
& 1.7 & 2.6 & 0.7 & 2.8 & 2.0 \\

& ReFusion (2019)~\cite{refusion} & D
& 28.9 & 46.3 & 17.5 & 17.9 & 27.7
& 9.9 & 10.4 & 1.7 & 11.0 & 8.3 \\

& ORB-SLAM3 (2021)~\cite{orbslam3} & S
& 70.7 & 77.9 & 5.8 & 3.1 & 39.4
& 28.1 & 30.5 & 2.0 & 2.6 & 15.8 \\

\midrule

\multirow{3}{*}{\parbox{1.6cm}{\centering NeRF-based\\ SLAM methods}}
& NICE-SLAM (2022)~\cite{nice} & D
& 24.5 & 53.6 & 24.4 & 8.3 & 27.7
& 113.8 & X & 88.2 & 45.0 & -- \\

& ESLAM (2023)~\cite{eslam} & D
& 48.0 & 51.4 & 22.6 & 17.7 & 34.9
& 45.7 & 60.8 & 93.6 & 3.6 & 50.9 \\

& RoDyn-SLAM (2024)~\cite{rodyn} & D
& 14.5 & 13.8 & 7.9 & 12.6 & 12.2
& 8.3 & 5.6 & 1.7 & 4.4 & 5.0 \\

\midrule

\multirow{6}{*}{\parbox{1.6cm}{\centering 3DGS-based\\ SLAM methods}}
& Gaussian-SLAM (2023)~\cite{gaussianslam} & D
& 109.2 & 118.7 & 65.2 & 31.7 & 81.2
& 133.7 & 80.7 & 19.1 & 5.4 & 59.7 \\

& SplaTAM (2024)~\cite{splatam} & D
& 138.4 & 126.3 & 35.8 & 22.0 & 80.6
& 149.2 & 157.8 & 85.3 & 14.0 & 101.6 \\

& GassiDy (2025)~\cite{GassiDy} & D
& 10.3 & 13.0 & 2.6 & 5.4 & 7.8
& 3.5 & 3.7 & 0.6 & 2.4 & 2.6 \\

& WildGS-SLAM (2025)~\cite{wildgs} & D
& 3.9 & 3.1 & 2.8 & 2.5 & 3.1
& 1.3 & 1.6 & \textbf{0.4} & 2.0 & 1.3 \\

& DROID-W (2026)~\cite{droidw} & D
& \textbf{3.3} & \textbf{2.9} & 2.6 & 2.3 & 2.8
& 1.2 & 1.6 & 0.5 & \textbf{1.4} & \textbf{1.2} \\

& \textbf{DPGS-SLAM (Ours)} & D
& 3.5 & 3.1 & \textbf{2.0} & \textbf{2.1} & \textbf{2.7}
& \textbf{1.1} & \textbf{1.5} & \textbf{0.4} & 1.8 & \textbf{1.2} \\

\bottomrule
\end{tabular}
\end{lrbox}

\begin{minipage}{\wd\trackingtabbox}
\usebox{\trackingtabbox}

\par\vspace{3pt}
\scriptsize
\raggedright
\textit{Notation:}
``--'' denotes absence of mention, and ``X'' indicates tracking failures.
\textbf{S} and \textbf{D} denote sparse and dense SLAM, respectively.
\end{minipage}
\end{table*}


\begin{figure*}[t]
\centering
\scriptsize

\setlength{\tabcolsep}{2pt}
\renewcommand{\arraystretch}{0.5}

\begin{tabular}{ccccccc}
& Input \textbf{GT} (ground-truth)& \textbf{SplaTAM} (2024)~\cite{splatam} & \textbf{MonoGS} (2024)~\cite{monogs} & \textbf{WildGS-SLAM} (2025)~\cite{wildgs}& \textbf{DROID-W} (2026)~\cite{droidw}  & \textbf{Ours} \\
\\[-2pt]

\raisebox{1\height}{\rotatebox{90}{wk.xyz}} &
\includegraphics[width=0.155\linewidth]{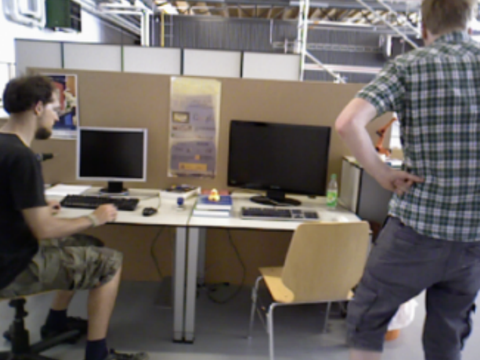} &
\includegraphics[width=0.155\linewidth]{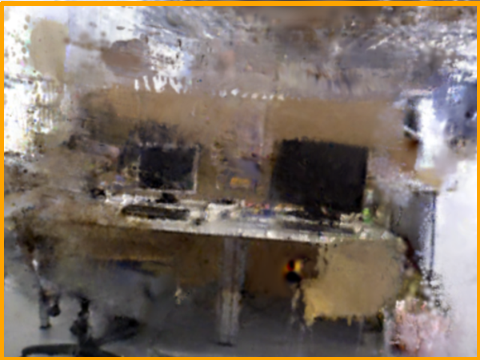} &
\includegraphics[width=0.155\linewidth]{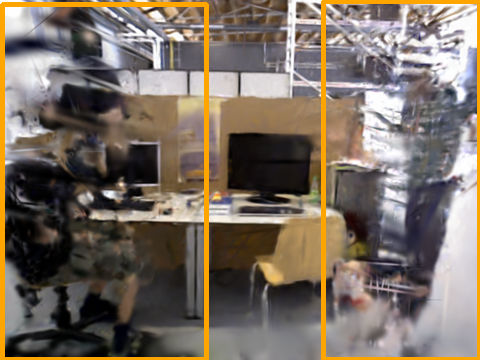} &
\includegraphics[width=0.155\linewidth]{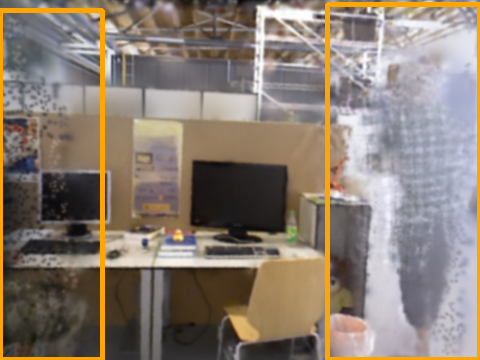} &
\includegraphics[width=0.155\linewidth]{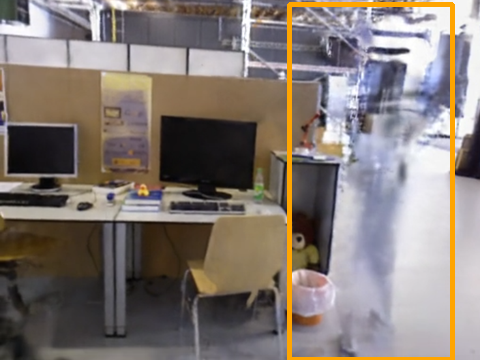} &
\includegraphics[width=0.155\linewidth]{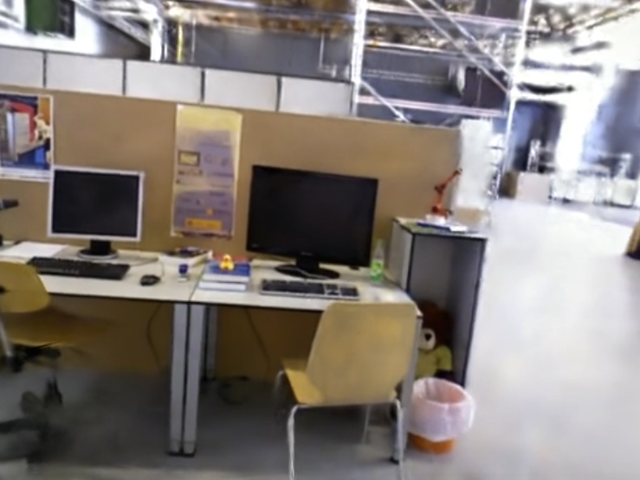}
\\

\raisebox{0.7\height}{\rotatebox{90}{\textbf{trajectory}}} &
\includegraphics[width=0.155\linewidth]{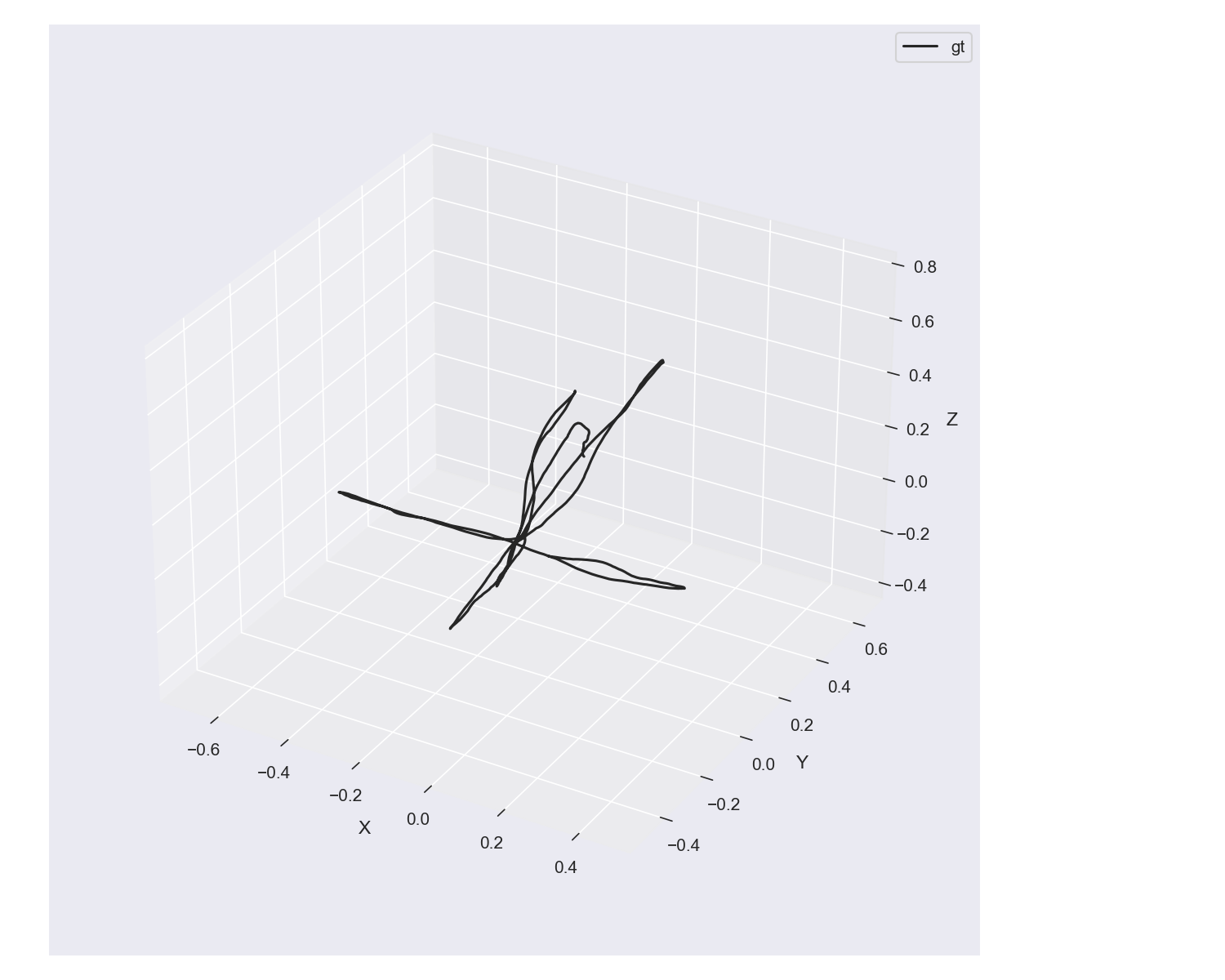} &
\includegraphics[width=0.155\linewidth]{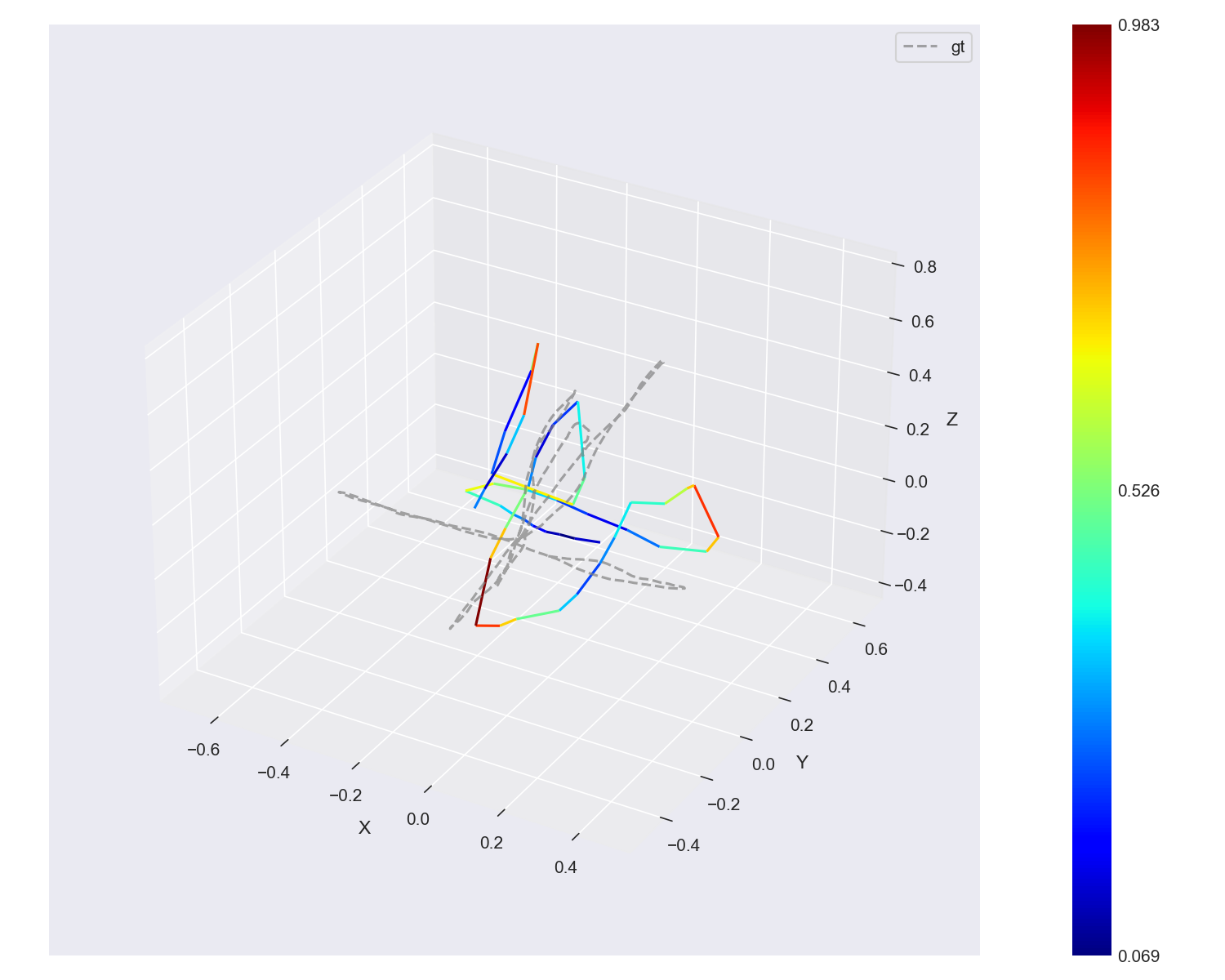} &
\includegraphics[width=0.155\linewidth]{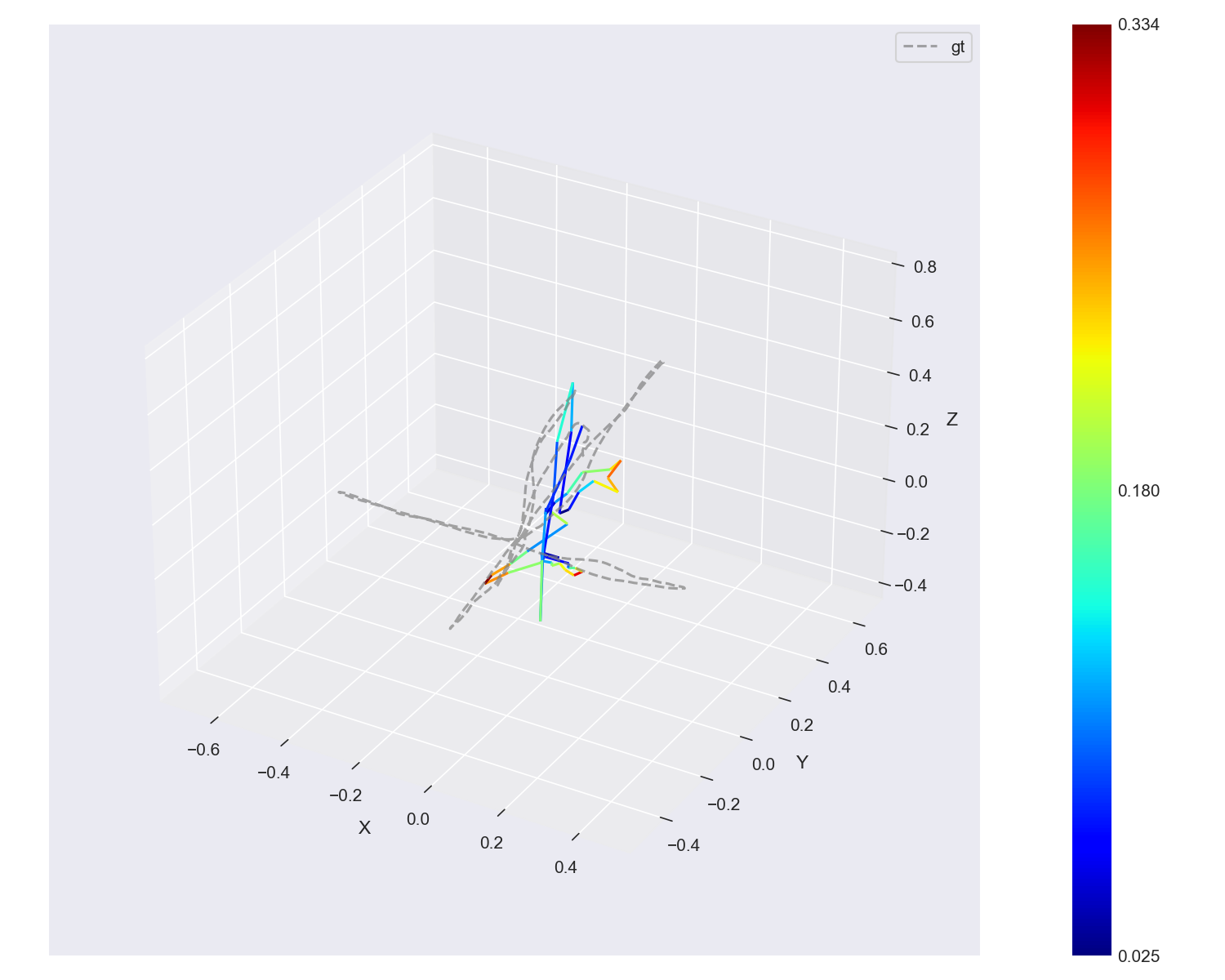} &
\includegraphics[width=0.155\linewidth]{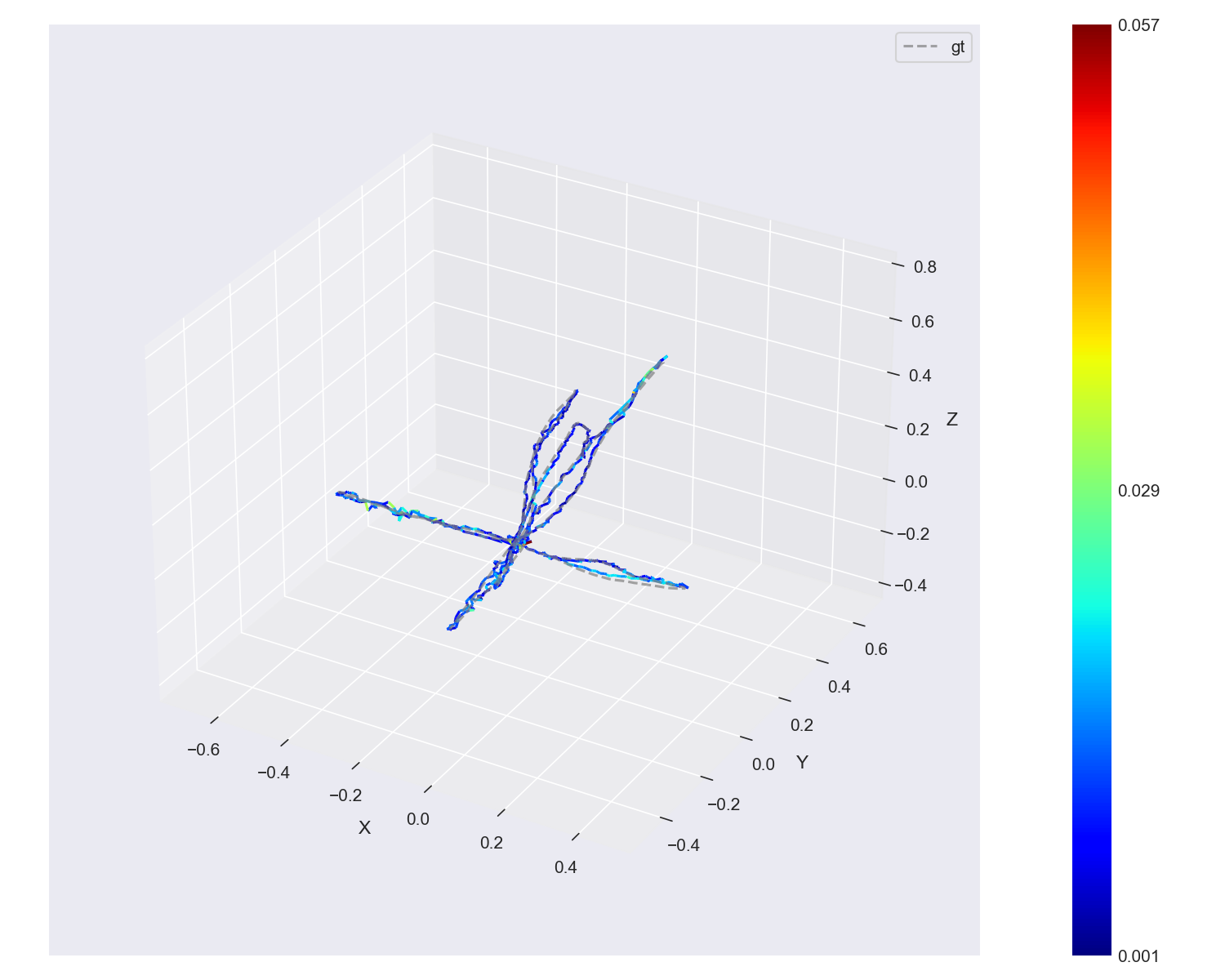} &
\includegraphics[width=0.155\linewidth]{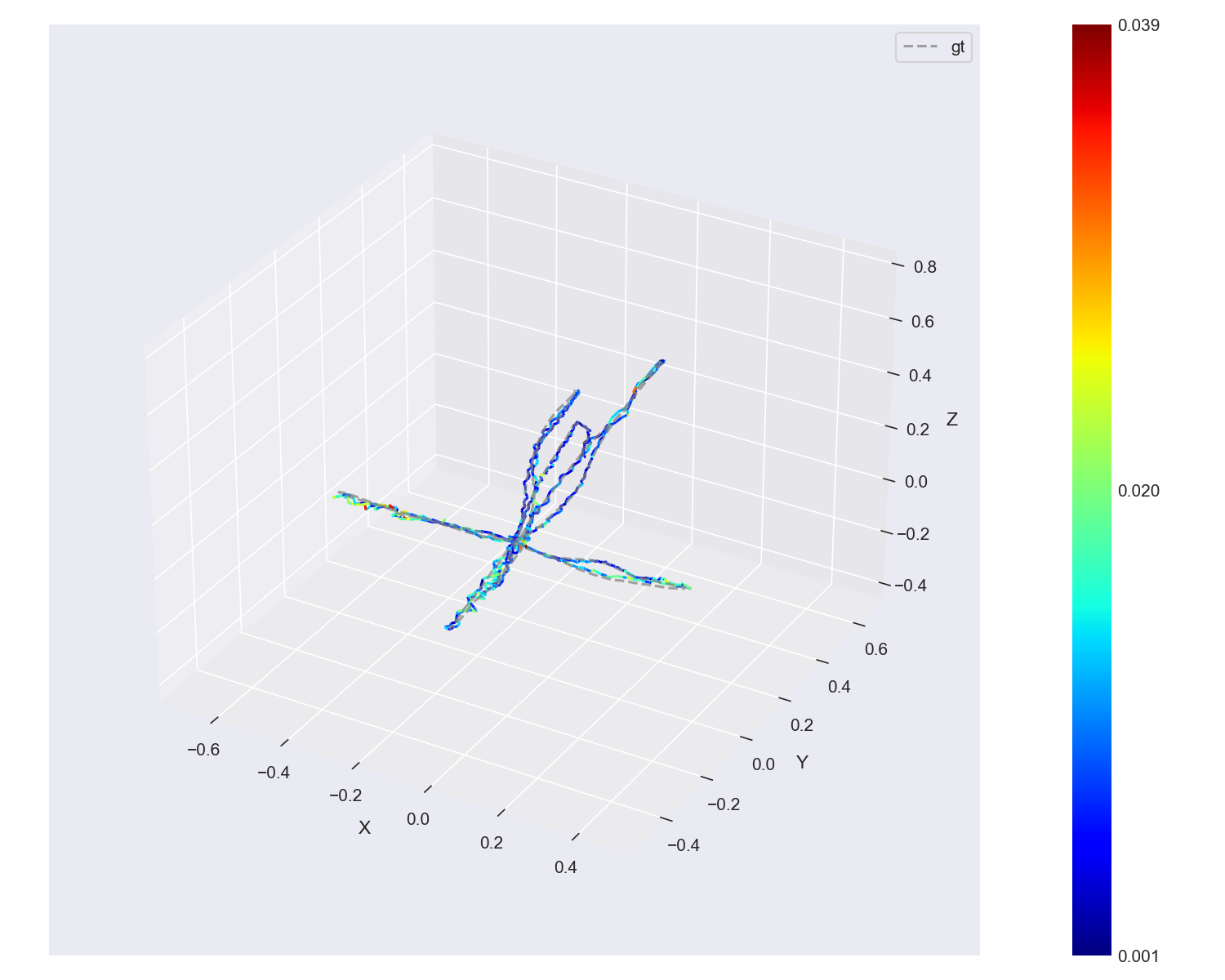} &
\includegraphics[width=0.155\linewidth]{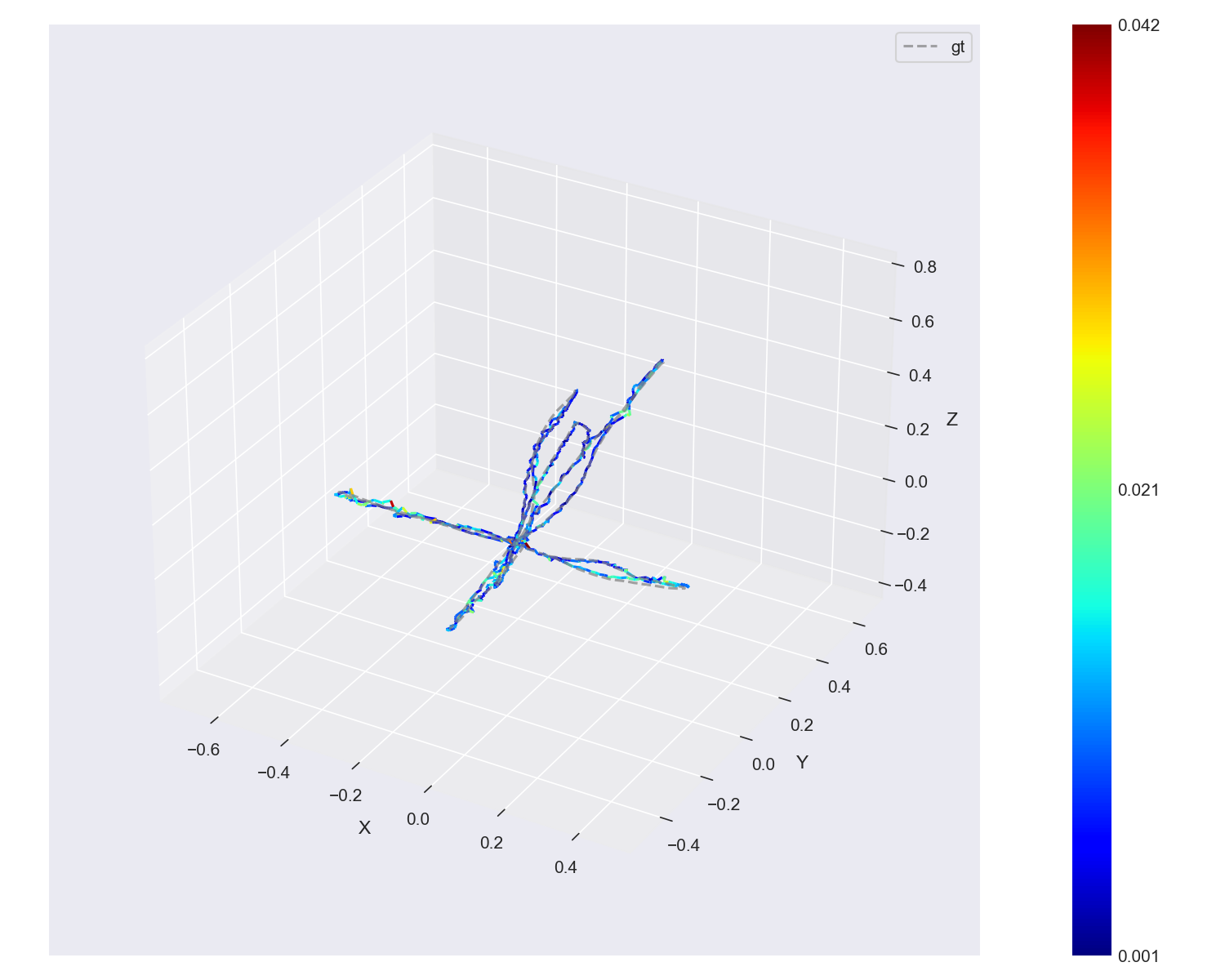}
\\

\raisebox{1.5\height}{\rotatebox{90}{ps.tr2}} &
\includegraphics[width=0.155\linewidth]{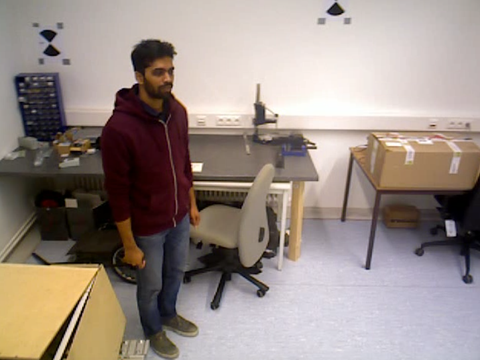} &
\includegraphics[width=0.155\linewidth]{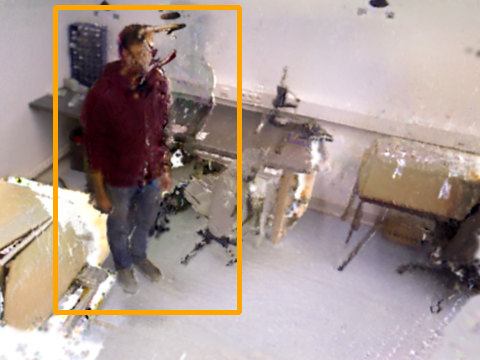} &
\includegraphics[width=0.155\linewidth]{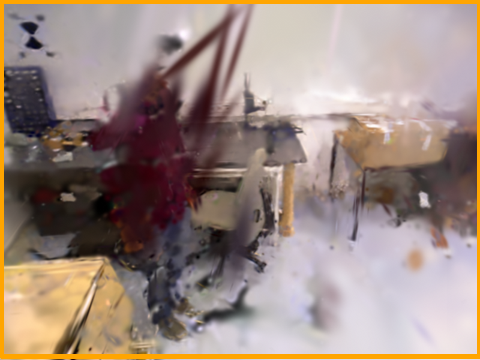} &
\includegraphics[width=0.155\linewidth]{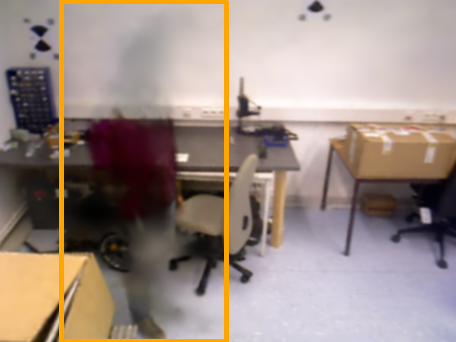} &
\includegraphics[width=0.155\linewidth]{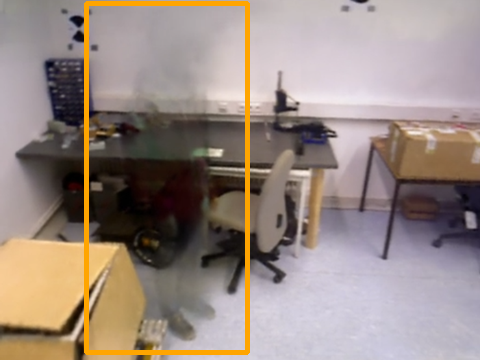} &
\includegraphics[width=0.155\linewidth]{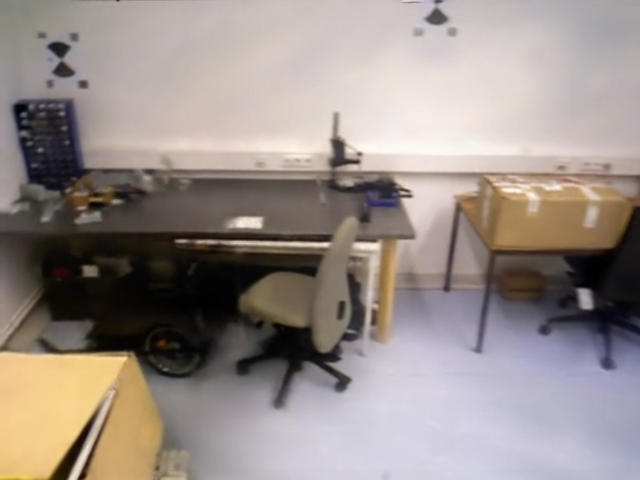}
\\

\raisebox{0.7\height}{\rotatebox{90}{\textbf{trajectory}}} &
\includegraphics[width=0.155\linewidth]{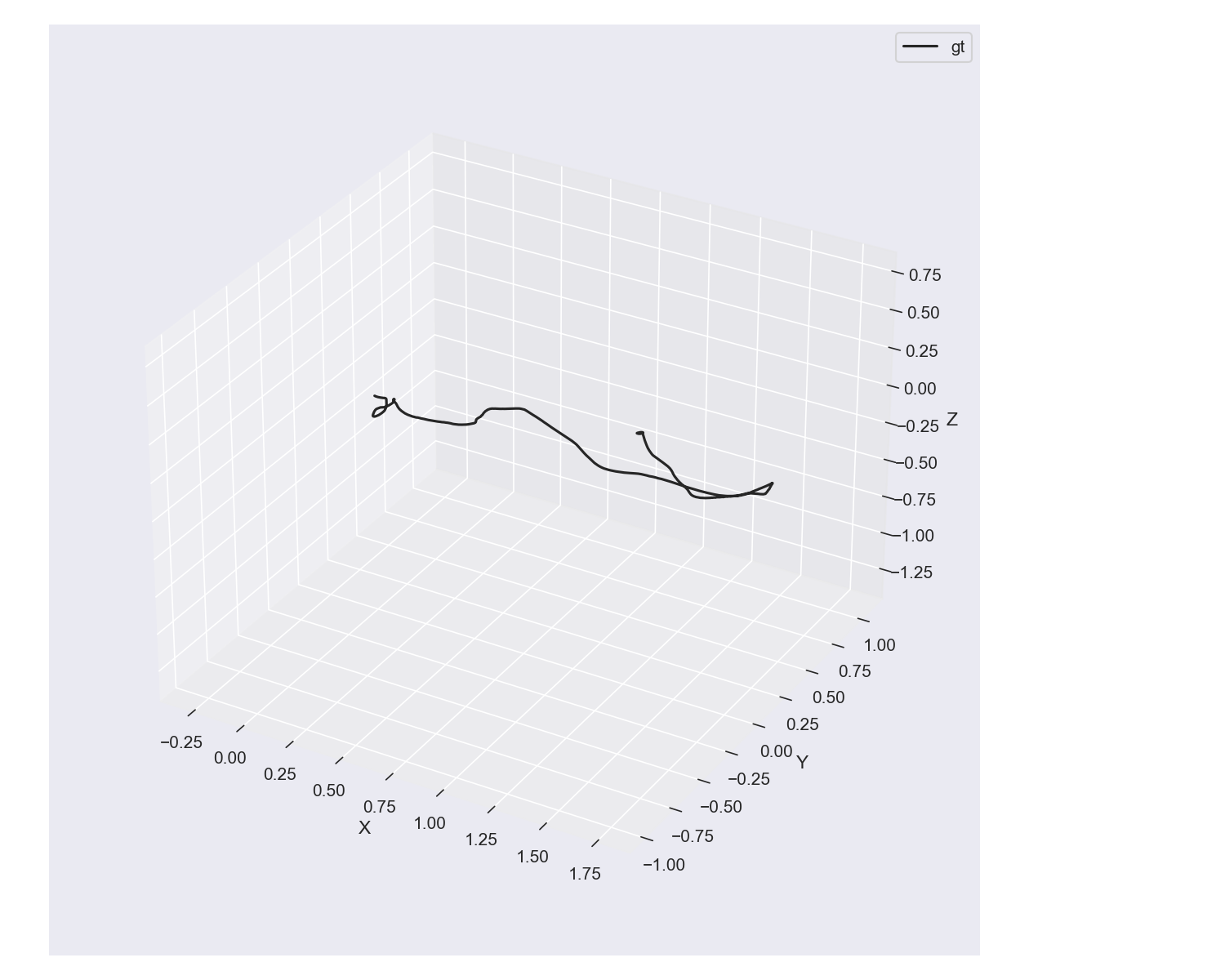} &
\includegraphics[width=0.155\linewidth]{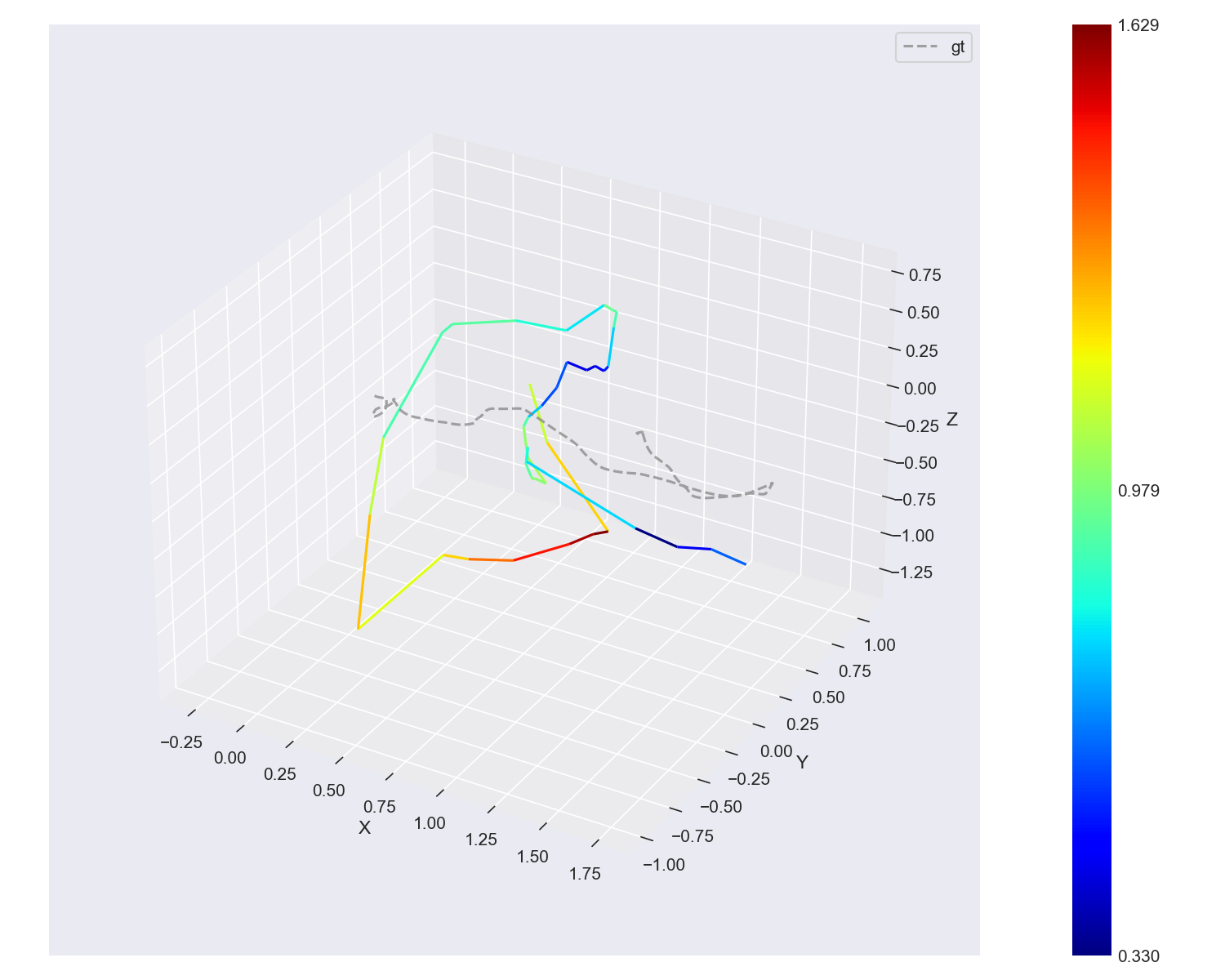} &
\includegraphics[width=0.155\linewidth]{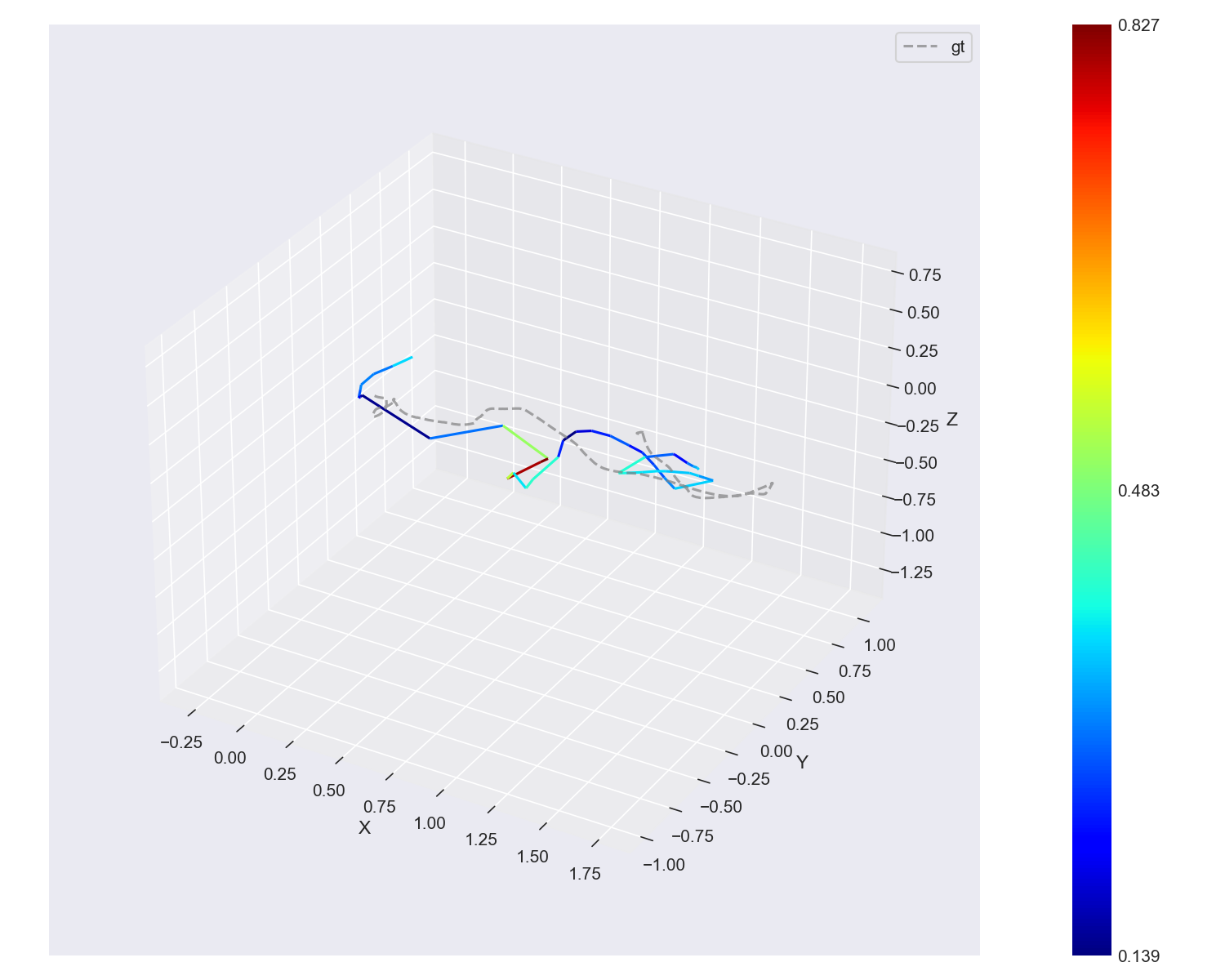} &
\includegraphics[width=0.155\linewidth]{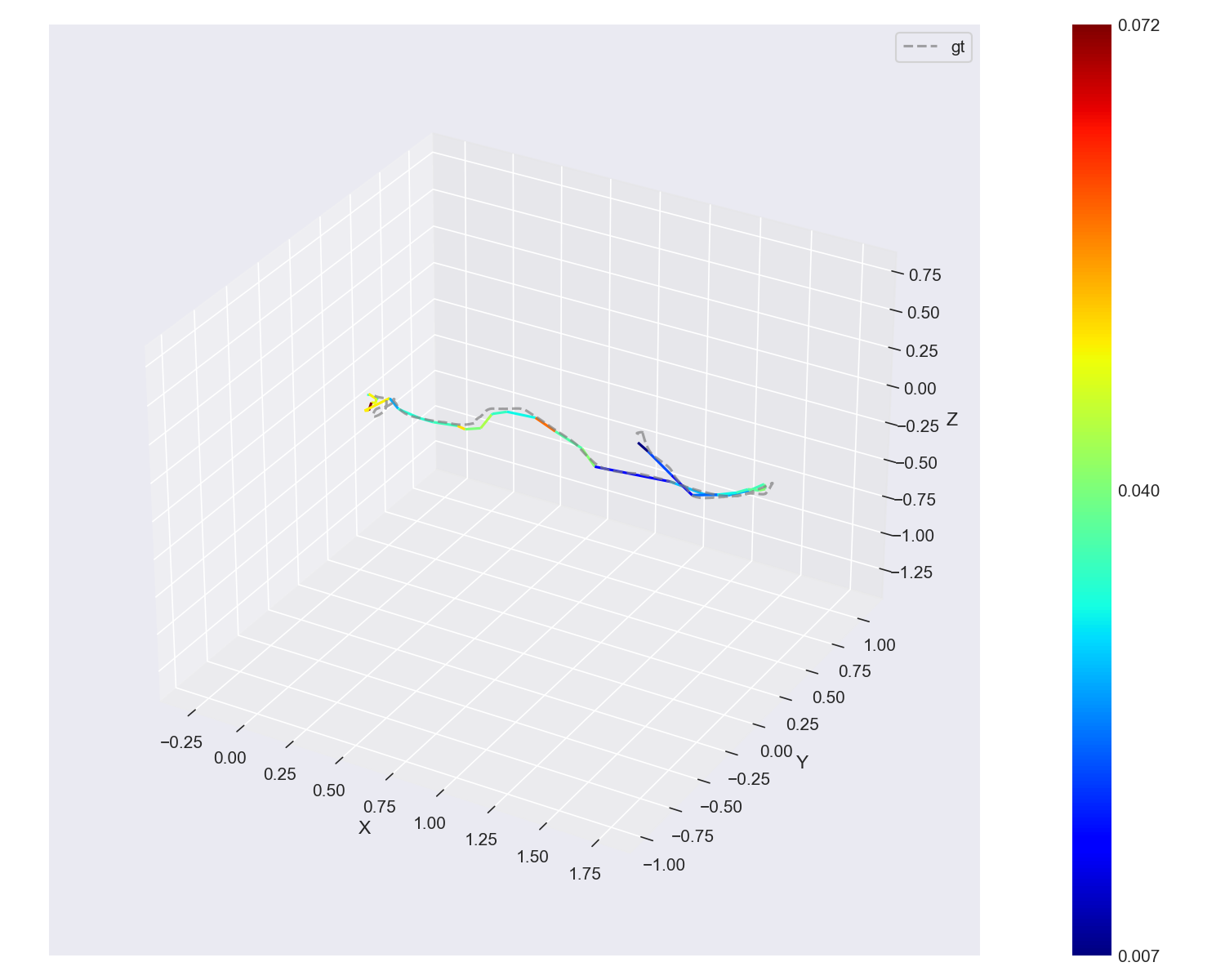} &
\includegraphics[width=0.155\linewidth]{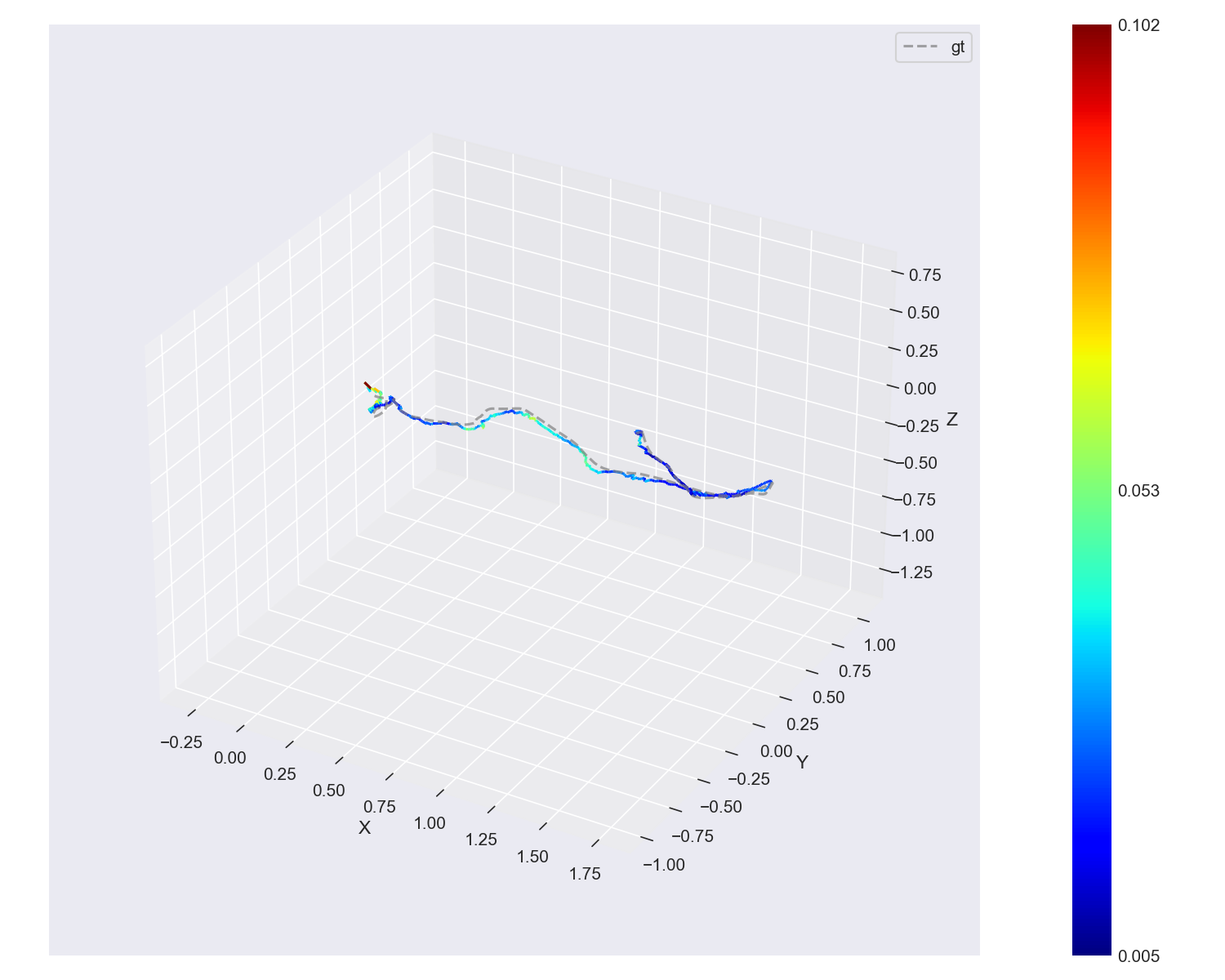} &
\includegraphics[width=0.155\linewidth]{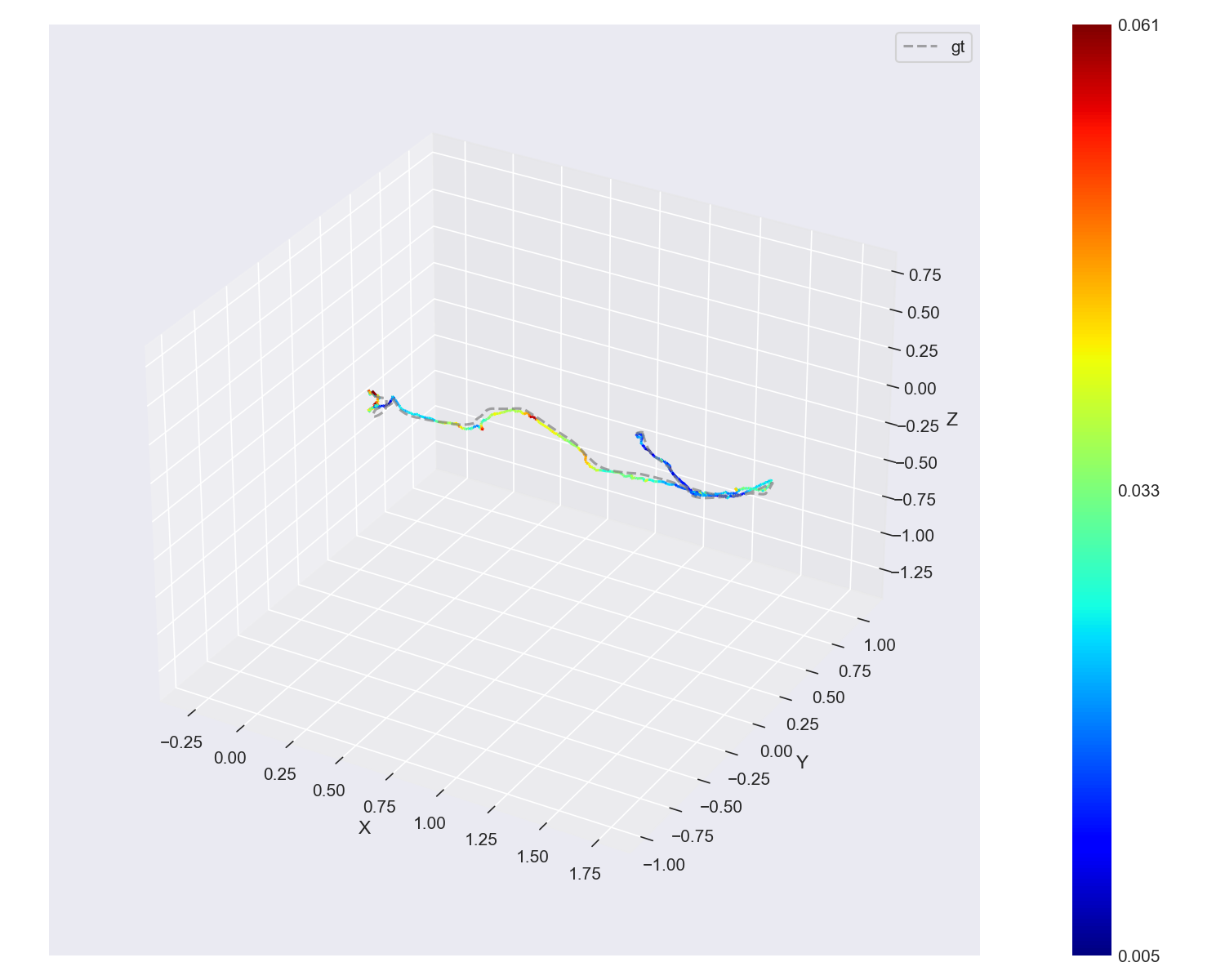}
\\

\end{tabular}

\caption{Visual comparison of the rendered images and corresponding camera trajectories on TUM RGB-D (wk.xyz) and BONN RGB-D (ps.tr2) datasets. 
Orange boxes highlight artifacts in baseline renderings, including ghosting, blurring, and local geometric distortions.
}
\label{fig:render_traj_compare}
\end{figure*}


\begin{figure}[t]
    \centering
    \scriptsize
    \setlength{\tabcolsep}{2pt}
    \renewcommand{\arraystretch}{0.9}

    \begin{tabular}{cccc}
    input RGB & \textbf{RoDyn-SLAM}~\cite{rodyn} & \textbf{DROID-W}~\cite{droidw} & \textbf{Ours} \\

    \includegraphics[width=0.235\columnwidth]{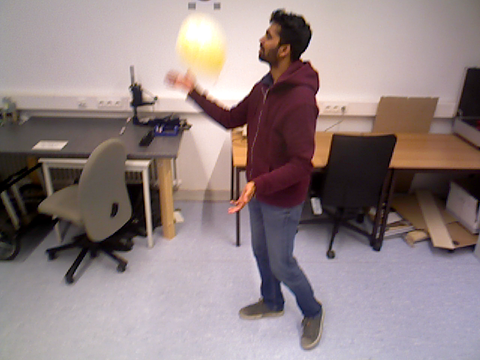} &
    \includegraphics[width=0.235\columnwidth]{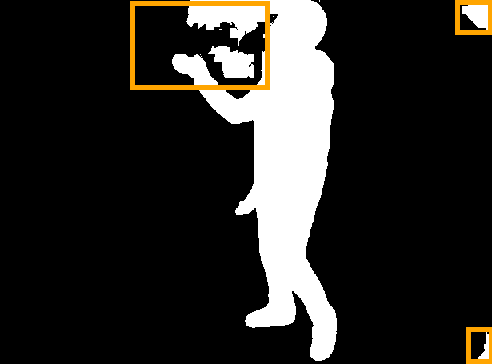} &
    \includegraphics[width=0.235\columnwidth]{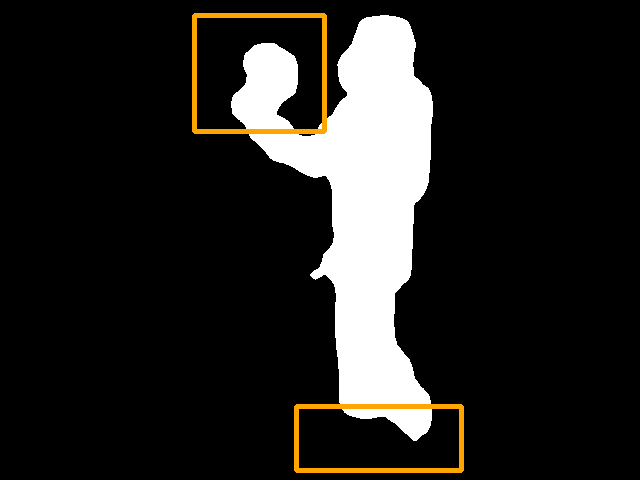} &
    \includegraphics[width=0.235\columnwidth]{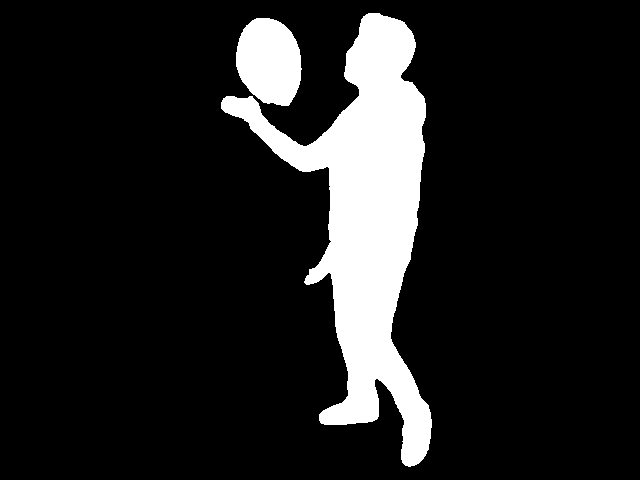} \\[-2pt]
    \\[-5pt]

    \includegraphics[width=0.235\columnwidth]{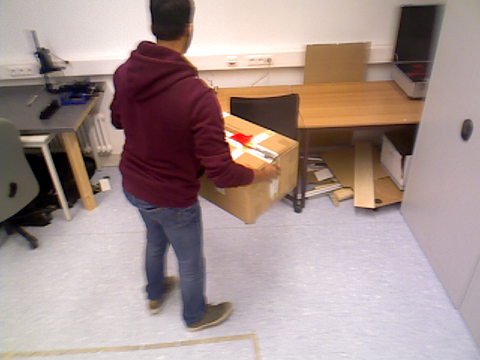} &
    \includegraphics[width=0.235\columnwidth]{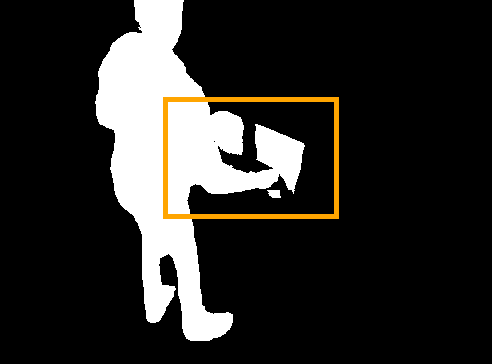} &
    \includegraphics[width=0.235\columnwidth]{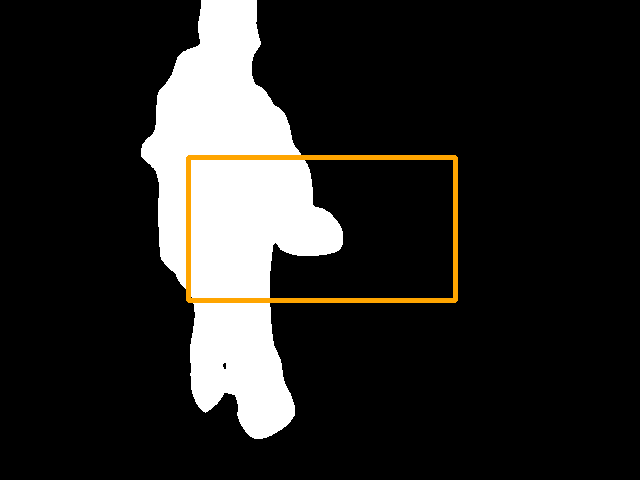} &
    \includegraphics[width=0.235\columnwidth]{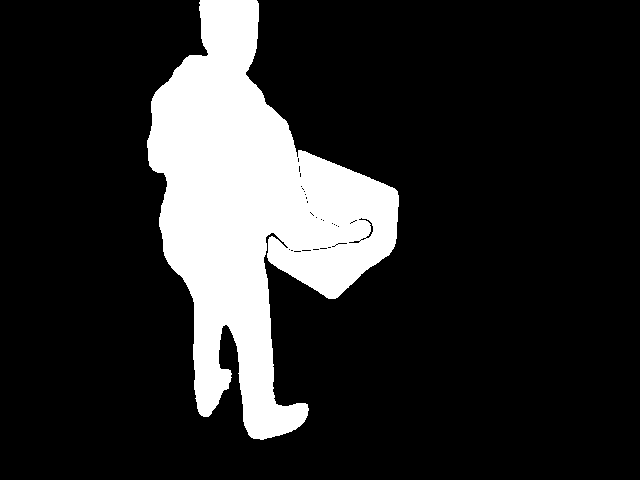} \\[-2pt]

    \end{tabular}

    \caption{
        Qualitative comparison of dynamic-region estimates on BONN RGB-D dataset.
        From left to right, each row shows the input RGB frame, the RoDyn-SLAM motion mask, the DROID-W dynamic uncertainty map, and the DPGS-SLAM motion mask.
        Orange boxes highlight regions with incorrect motion segmentation (e.g., false positives or missed dynamic areas).
    }
    \label{fig:mask_comparison}
\end{figure}

\section{Experiments and Results} \label{sec:exp}

\subsection{Experimental Setup}

\subsubsection{Datasets}
We evaluated our DPGS-SLAM on two widely used dynamic RGB-D benchmarks: 
(a) \textbf{TUM RGB-D} (2012)~\cite{tum} dataset and (b) \textbf{BONN RGB-D} dynamic (2019)~\cite{refusion} dataset. 
Both datasets capture indoor scenes using handheld RGB-D cameras and provide ground-truth camera trajectories. 
Following prior work on dynamic SLAM such as GassiDy (2025)~\cite{GassiDy} and WildGS-SLAM (2025)~\cite{wildgs}, we selected sequences involving human motion and object interactions to evaluate robustness in dynamic scenes.

We additionally evaluated static-background reconstruction on three sequences from the \textbf{StaDy4D} (2026)~\cite{stady4d} dataset.
StaDy4D is a synthetic dataset that provides paired static and dynamic RGB sequences rendered along identical camera trajectories.
The dynamic sequences were used as SLAM inputs, while the corresponding static sequences provided ground-truth (GT) references for evaluating the reconstructed background and were not involved in tracking or mapping optimization.

For qualitative ablation comparisons, we used the \textit{wandering} sequence from WildGS-SLAM iPhone (2025)~\cite{wildgs} and \textit{freiburg3\_sitting\_halfsphere} from TUM RGB-D~\cite{tum}, representing limited background observations under rapid viewpoint changes and prolonged human occlusion, respectively.

Across all evaluations, our DPGS-SLAM used only monocular RGB images and camera intrinsics as input; neither sensor nor ground-truth depth was used for tracking, mapping, or background completion. 
Ground-truth depth from StaDy4D was used only for evaluation, including the construction of regional masks and quantitative assessment of geometric accuracy.

For LoRA-based fine-tuning of the diffusion inpainting model for static-background completion (Section~\ref{sec:back}), we used indoor scene images sampled from \textbf{ScanNet} (2017)~\cite{scannet} and \textbf{ScanNet++} (2023)~\cite{scannet++} datasets.
Synthetic occlusion masks were applied to these images to construct the masked-image training pairs described in~\eqref{eq:diff-masked-input}--\eqref{eq:diffusion}.


\subsubsection{Metrics}
Camera tracking accuracy was evaluated using the ATE (absolute trajectory error), reported as RMSE (root mean squared error) in centimeters. Following WildGS-SLAM~\cite{wildgs} and DROID-W~\cite{droidw}, we evaluated the complete camera trajectory, including non-keyframes, after aligning the estimated trajectory with the ground-truth using \texttt{evo} (2017)~\cite{evo} with $\mathrm{Sim}(3)$ Umeyama alignment~\cite{umeyama}.

To assess reconstruction quality under dynamic scenes, we report standard photometric rendering metrics commonly adopted in 3DGS-SLAM (like GassiDy~\cite{GassiDy} and WildGS-SLAM~\cite{wildgs}), including PSNR (peak signal-to-noise ratio), SSIM (structural similarity index)~\cite{ssim}, and LPIPS (learned perceptual image patch similarity)~\cite{lpips}. On StaDy4D dataset, full-frame renderings were compared with paired static RGB frames at matching time steps, providing a background-reconstruction reference that is independent of the dynamic input frames.


\subsubsection{Baselines}
We compared our DPGS-SLAM against the state-of-the-art methods from different SLAM paradigms: 
\textbf{Traditional} ones (DynaSLAM (2018)~\cite{dynaslam}, ReFusion (2019)~\cite{refusion}, ORB-SLAM3 (2021)~\cite{orbslam3}) including both sparse and dense SLAM methods, 
\textbf{Neural implicit} ones (NICE-SLAM (2022)~\cite{nice}, ESLAM (2023)~\cite{eslam}, RoDyn-SLAM (2024)~\cite{rodyn}) as dense but NeRF-based SLAM,
and \textbf{3DGS-based} ones (Gaussian-SLAM (2023)~\cite{gaussianslam}, MonoGS (2024)~\cite{monogs}, SplaTAM (2024)~\cite{splatam}, as well as dynamic-oriented methods GassiDy (2025)~\cite{GassiDy}, WildGS-SLAM (2025)~\cite{wildgs} and DROID-W (2026)~\cite{droidw}) as dense SLAM similar to ours. 
This selection supports a comparison of both tracking robustness and reconstruction quality in dynamic environments. 

For the tracking comparison in Table~\ref{tab:tracking_results}, baseline results were collected from established evaluations on the TUM RGB-D and BONN RGB-D benchmarks, as reported in RoDyn-SLAM~\cite{rodyn}, GassiDy~\cite{GassiDy}, and DROID-W~\cite{droidw}. The corresponding input modalities and method variants followed these published benchmarks, with average ATE RMSE values recalculated over the four selected sequences of each dataset. For StaDy4D, MonoGS~\cite{monogs}, WildGS-SLAM~\cite{wildgs}, and DROID-W~\cite{droidw} were evaluated using their public implementations with identical dynamic RGB inputs and mapping optimization settings. DROID-W's optional Gaussian mapping module was enabled (\texttt{mapping.enable: True}) for reconstruction evaluation.


\subsubsection{Implementation Details}
All experiments were conducted on a workstation equipped with an NVIDIA RTX A6000 GPU.
Motion discovery was performed every $S=10$ frames using $\Delta=10$ consecutive pairs of forward and backward optical flows, with an object-association IoU threshold of $0.5$.
We used a sliding window of $|\mathcal{W}_{\mathrm c}|=10$ keyframes together with two randomly sampled past keyframes.
Observation-supported Gaussian primitives were initialized by randomly sampling a fraction $\rho=0.0625$ of valid static pixels.
Background completion was triggered when $a_t \geq \tau_{\mathrm{comp}}=0.03$, with $\tau_{\mathrm{cov}}=0.5$ used to identify insufficient map coverage, prioritizing sizable unsupported regions while limiting requests for small residual gaps.
The replacement tolerance $\tau_{\mathrm{replace}}=0.05\,\mathrm{m}$ allows a small margin behind the predicted static surface to accommodate depth discrepancies while retaining generated geometry farther behind it.
The loss weights in~\eqref{eq:lossTrack} and~\eqref{eq:lossMap} were set to
$\lambda_{\mathrm{pho}}=0.9$,
$\lambda_{\mathrm{geo}}=0.1$, and
$\lambda_{\mathrm{iso}}=10$.
Camera pose optimization used Adam~\cite{adam}, with learning rates of 0.003 for rotation and 0.001 for translation, and $N_{\mathrm{track}}=100$ iterations per frame.
Map optimization used 150 iterations per update.
For LoRA-based fine-tuning, we used a LoRA rank of 8 and a learning rate of $10^{-4}$, while diffusion inference used 30 denoising steps with a guidance scale of 7.5.
The remaining Gaussian-parameter optimization settings followed a previous study~\cite{3dgs}.
The numerical settings were selected empirically through preliminary experiments and, unless otherwise specified, kept fixed across all sequences.

\subsection{Camera Tracking Accuracy Analysis}

We evaluated the camera tracking accuracy on dynamic scenes from the BONN RGB-D dynamic and TUM RGB-D datasets using the ATE RMSE metric.

\subsubsection{Quantitative Results} 

As shown in Table~\ref{tab:tracking_results}, our DPGS-SLAM achieved the lowest average ATE RMSE of 2.7\,cm on the four selected BONN RGB-D sequences. On TUM RGB-D, it achieved an average ATE RMSE of 1.2\,cm, matching DROID-W at the reported precision. This tracking performance is supported by motion-mask-based exclusion of dynamic observations and photometric--geometric pose optimization, as further examined in Table~\ref{tab:ablation_tracking}. Camera tracking uses only observation-supported Gaussian primitives, while diffusion-generated primitives remain excluded from pose optimization.

\subsubsection{Trajectory Analysis}

As shown in Fig.~\ref{fig:render_traj_compare}, our DPGS-SLAM maintained closer alignment with the ground-truth trajectory and exhibited reduced drift in the illustrated sequences, compared with SplaTAM and MonoGS.
On the \textit{wk.xyz} sequence of TUM RGB-D, DPGS-SLAM and DROID-W exhibited similar trajectory alignment, while WildGS-SLAM showed larger local deviations.
On the \textit{ps.tr2} sequence of BONN RGB-D, all three methods closely followed the ground-truth trajectory, with minor local deviations.

\subsection{Map and Motion Mask Quality Analysis}

\subsubsection{Dynamic-Scene Rendering Quality on TUM RGB-D and BONN RGB-D}

Fig.~\ref{fig:render_traj_compare} also presents qualitative comparisons of rendered images on TUM RGB-D and BONN RGB-D datasets.
Compared with the evaluated baselines, our DPGS-SLAM produced reconstructions with fewer ghosting artifacts in regions affected by dynamic objects.

\subsubsection{Motion Mask Quality on BONN RGB-D}

We qualitatively compared the motion masks from RoDyn-SLAM~\cite{rodyn} with the dynamic uncertainty maps from DROID-W~\cite{droidw} and the motion masks from DPGS-SLAM on the same RGB frames from two BONN RGB-D sequences (\textit{mv\_box2} and \textit{balloon}).
As shown in Fig.~\ref{fig:mask_comparison}, the DPGS-SLAM masks aligned more closely with the visible moving objects in the illustrated frames.

\subsubsection{Static-Background Reconstruction on StaDy4D}

To quantitatively assess static-background reconstruction under dynamic occlusions, we conducted experiments on three sequences from StaDy4D benchmark~\cite{stady4d}: \textit{scene\_T07\_022\_06\_cctv}, \textit{scene\_T10\_024\_00\_car}, and \textit{scene\_T03\_016\_07\_pedestrian}.
We compared our DPGS-SLAM with MonoGS~\cite{monogs}, WildGS-SLAM~\cite{wildgs}, and DROID-W~\cite{droidw}.
All methods used dynamic RGB sequences as input, while the paired static RGB frames were used only as ground-truth references for reconstruction evaluation. 
For the rendering comparisons involving DROID-W, we enabled its optional Gaussian splatting mapping module (\texttt{mapping.enable: True}) in the official implementation. All evaluated methods used the same number of mapping optimization iterations.

As shown in Table~\ref{tab:stady4d_reconstruction}, our DPGS-SLAM achieved the highest PSNR and SSIM and the lowest LPIPS among the compared methods across all three evaluated sequences.
Compared with WildGS-SLAM, DPGS-SLAM improved the average PSNR by 4.60\,dB. The average SSIM increased by 7.97 percentage points, while the average LPIPS ($\times 100$) decreased by 2.97. These results mean better photometric, structural, and perceptual agreement with the static-background references.

\begin{figure*}[t]
\centering
\scriptsize

\setlength{\tabcolsep}{1.5pt}
\renewcommand{\arraystretch}{1.0}

\begin{tabular}{
@{}
>{\centering\arraybackslash}m{8mm}
*{6}{>{\centering\arraybackslash}m{0.15\textwidth}}
@{}
}

& \textbf{Dynamic Input}
& \textbf{Static GT}
& \textbf{MonoGS} (2024)~\cite{monogs}
& \textbf{WildGS} (2025)~\cite{wildgs}
& \textbf{DROID-W} (2026)~\cite{droidw}
& \textbf{Ours} \\[2pt]


\rotatebox{90}{\textbf{CCTV}} &
\includegraphics[width=\linewidth]{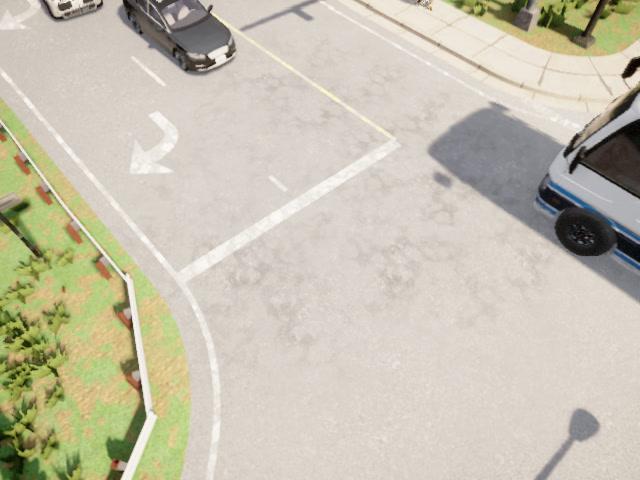} &
\includegraphics[width=\linewidth]{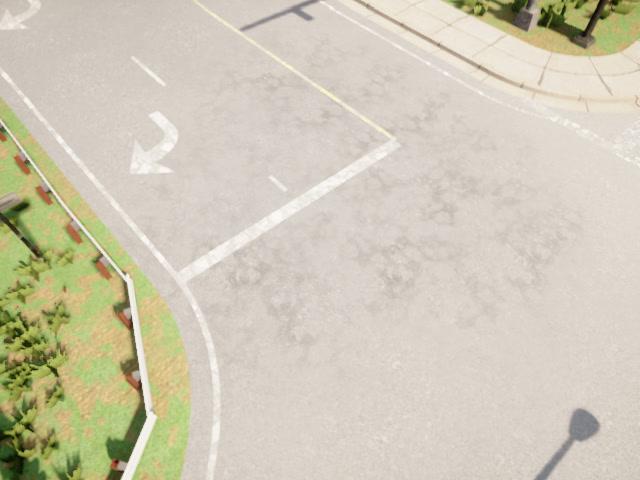} &
\includegraphics[width=\linewidth]{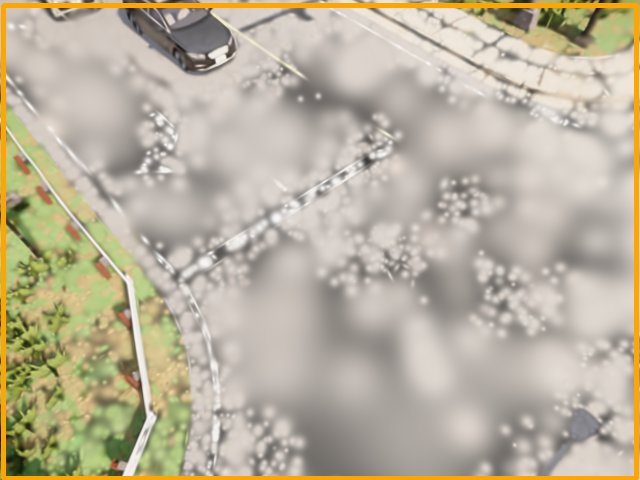} &
\includegraphics[width=\linewidth]{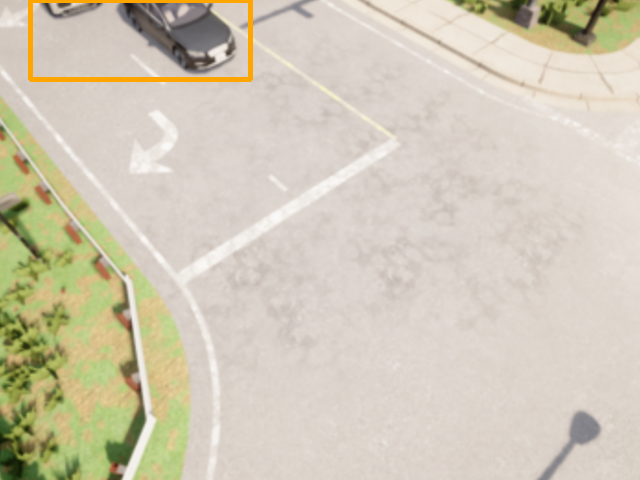} &
\includegraphics[width=\linewidth]{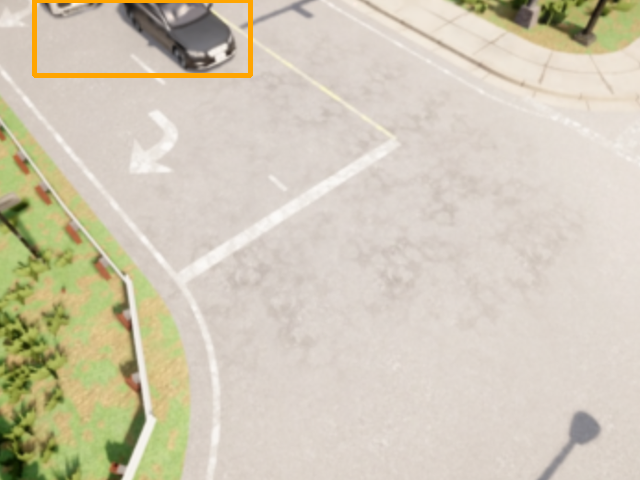} &
\includegraphics[width=\linewidth]{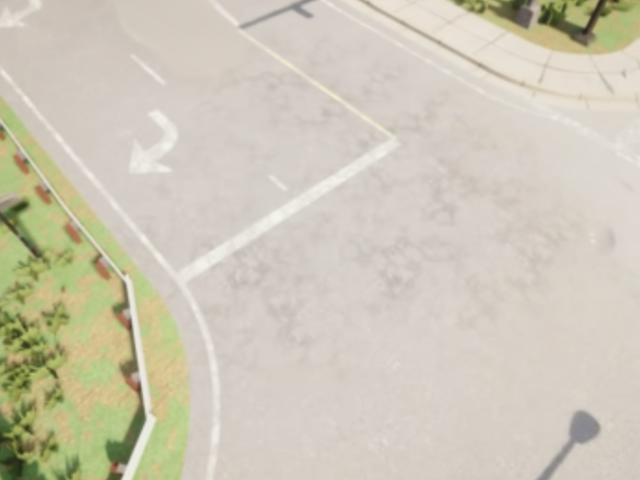}
\\[1pt]


\rotatebox{90}{\textbf{Car}} &
\includegraphics[width=\linewidth]{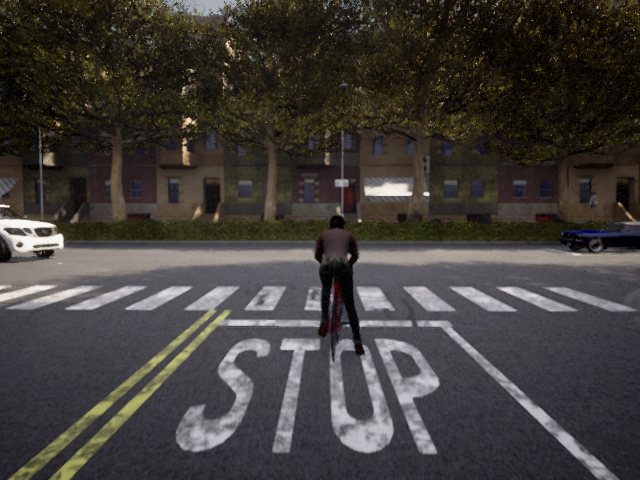} &
\includegraphics[width=\linewidth]{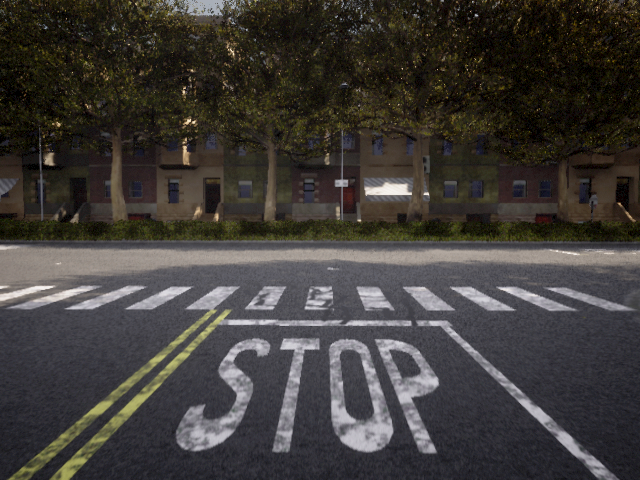} &
\includegraphics[width=\linewidth]{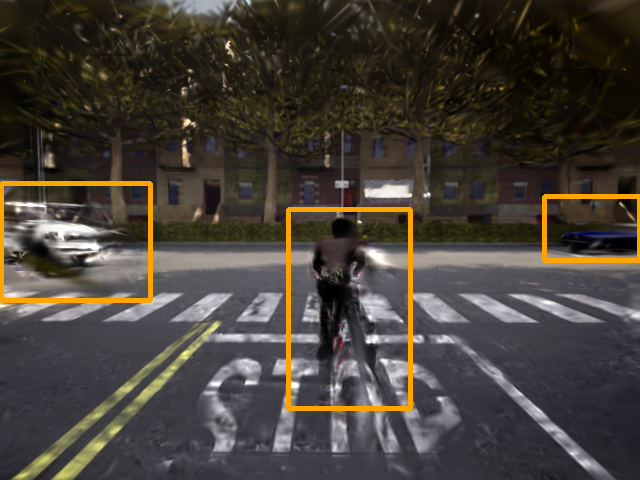} &
\includegraphics[width=\linewidth]{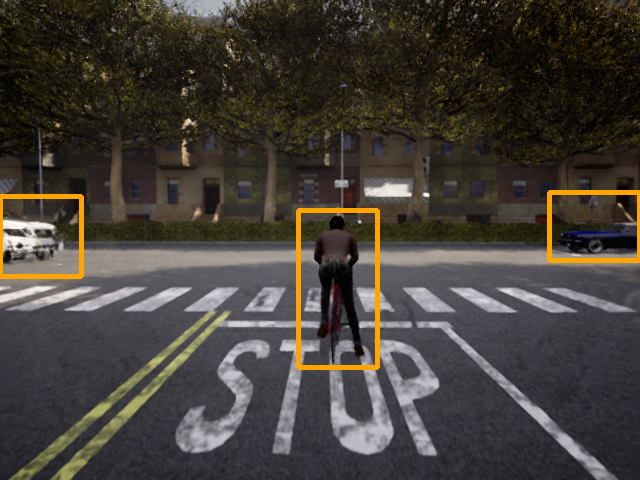} &
\includegraphics[width=\linewidth]{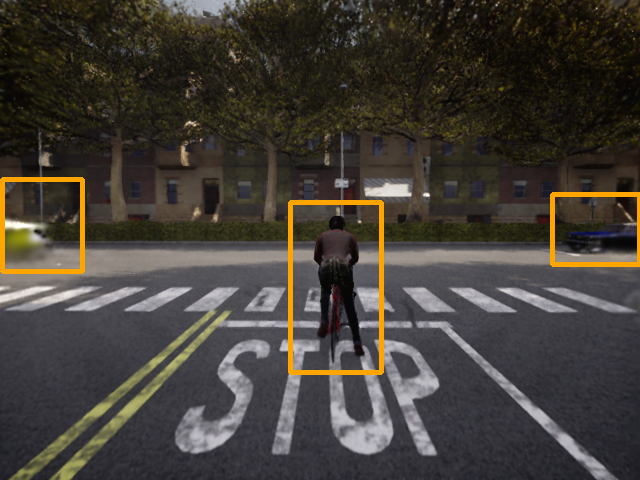} &
\includegraphics[width=\linewidth]{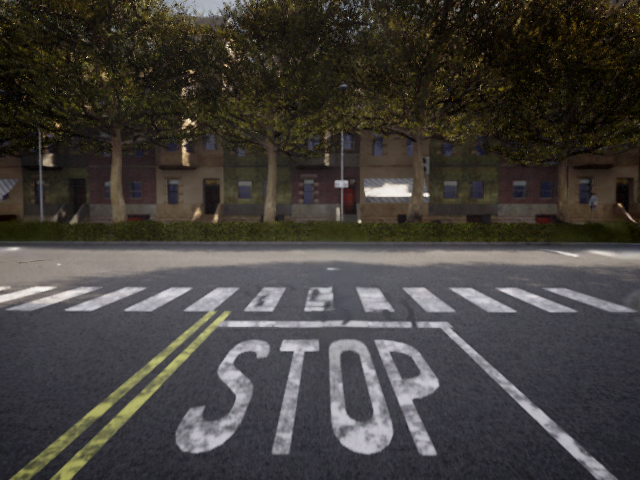}
\\[1pt]


\rotatebox{90}{\textbf{Pedestrian}} &
\includegraphics[width=\linewidth]{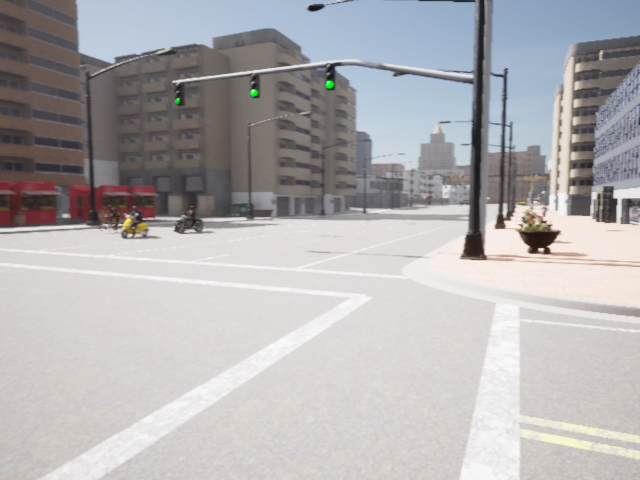} &
\includegraphics[width=\linewidth]{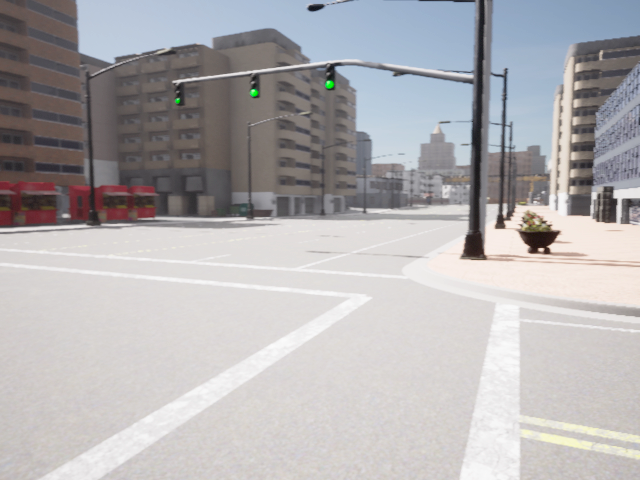} &
\includegraphics[width=\linewidth]{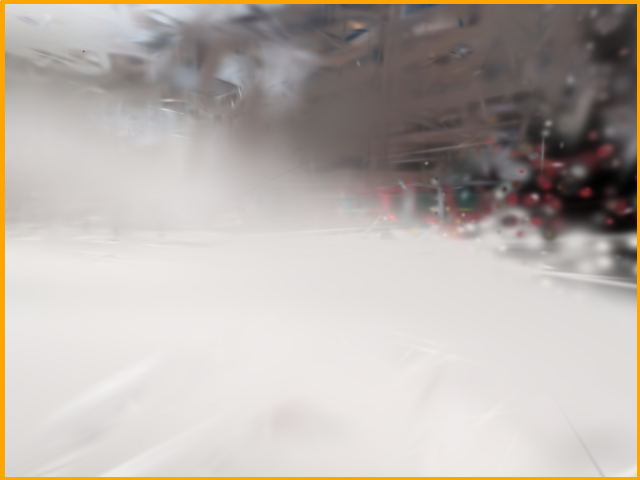} &
\includegraphics[width=\linewidth]{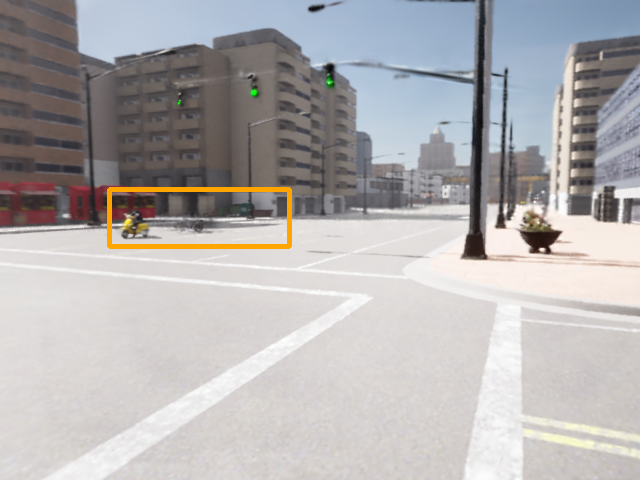} &
\includegraphics[width=\linewidth]{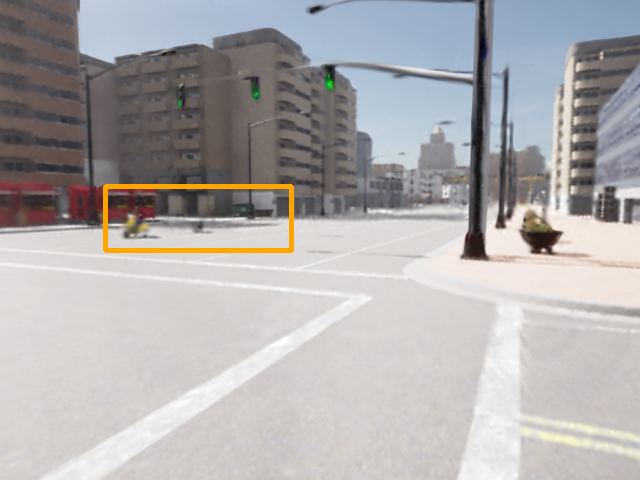} &
\includegraphics[width=\linewidth]{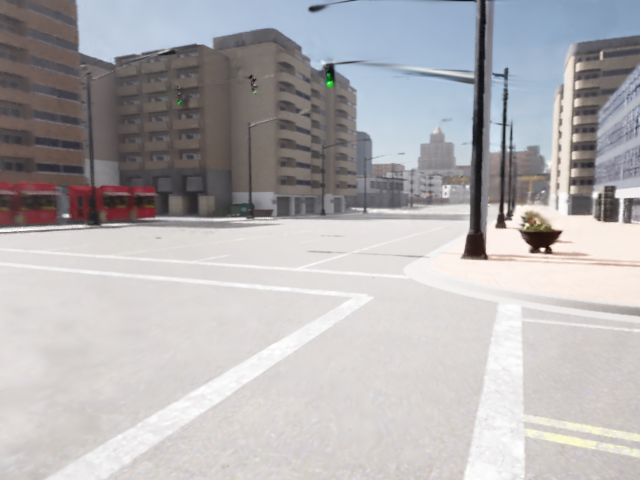}
\\

\end{tabular}

\caption{
Qualitative comparison of static-background reconstruction
on StaDy4D: \texttt{scene\_T07\_022\_06\_cctv} (top),
\texttt{scene\_T10\_024\_00\_car} (middle), and \texttt{scene\_T03\_016\_07\_pedestrian} (bottom).
All methods used dynamic RGB sequences for SLAM.
The first two columns show the dynamic input and corresponding static ground-truth, followed by the rendered images reconstructed using each method. Orange boxes highlight artifacts in baseline renderings, including ghosting, blurring, and residual moving objects.
}

\label{fig:stady4d_visual_comparison}
\end{figure*}

Qualitative comparisons are provided in Fig.~\ref{fig:stady4d_visual_comparison}, showing the dynamic inputs, paired static ground-truth, and reconstructed renderings for visual assessment of the occluded background regions.

\begin{table}[t]
\centering
\scriptsize
\caption{Static-background reconstruction quality on StaDy4D using dynamic RGB inputs and paired static ground-truth references.
}

\setlength{\tabcolsep}{3.5pt}
\label{tab:stady4d_reconstruction} 
\begin{tabular}{p{12mm}p{10mm}p{11mm}p{11mm}p{11mm}p{14mm}}
\toprule
\textbf{Metrics} & \textbf{Scene} & MonoGS (2024)~\cite{monogs} & WildGS (2025)~\cite{wildgs} & DROID-W (2026)~\cite{droidw} & DPGS-SLAM (Ours) \\
\midrule
\multirow{3}{*}{PSNR$\uparrow$ [dB]}
& T0722\_cctv & 16.6 & 18.9 & 21.8 & \textbf{25.2} \\
& T1024\_car & 16.3 & 19.9 & 19.7 & \textbf{24.5} \\
& T0316\_ped & 12.6 & 24.8 & 24.0 & \textbf{27.7} \\
\midrule
\multirow{3}{*}{SSIM$\uparrow$ [\%]}
& T0722\_cctv & 68.4 & 75.5 & 76.7 & \textbf{83.7} \\
& T1024\_car & 60.2 & 69.3 & 70.0 & \textbf{80.1} \\
& T0316\_ped & 56.1 & 84.0 & 83.2 & \textbf{88.9} \\
\midrule
\multirow{3}{*}{\makecell[c]{LPIPS$\downarrow$\\($\times 100$)}}
& T0722\_cctv & 37.5 & 29.7 & 30.0 & \textbf{28.4} \\
& T1024\_car & 45.5 & 26.2 & 26.7 & \textbf{20.1} \\
& T0316\_ped & 62.1 & 19.4 & 23.4 & \textbf{17.9} \\
\bottomrule
\end{tabular}
\\
\vspace{3pt}

\begin{minipage}{\columnwidth}
\scriptsize
\raggedright
\textit{Scene abbreviations:}
\texttt{T0722\_cctv}:
\texttt{scene\_T07\_022\_06\_cctv};
\texttt{T1024\_car}:
\texttt{scene\_T10\_024\_00\_car};
\texttt{T0316\_ped}:
\texttt{scene\_T03\_016\_07\_pedestrian}.
\textit{Notation:}
SSIM and LPIPS are multiplied by 100.
\end{minipage}
\end{table}

\subsubsection{Runtime Comparison on StaDy4D}
\label{sec:runtime_stady4d}

To evaluate computational efficiency, we compared the average per-frame tracking and mapping times of DPGS-SLAM with MonoGS~\cite{monogs}, WildGS-SLAM~\cite{wildgs}, and DROID-W~\cite{droidw} on the three StaDy4D sequences.
All methods were evaluated on the same NVIDIA RTX A6000 GPU using identical dynamic RGB inputs.
Tracking used 100 iterations per frame for DPGS-SLAM, at most 100 for MonoGS, and the original configurations for WildGS-SLAM and DROID-W.
For mapping, all methods used an average of 150 optimization iterations per keyframe, with DROID-W's Gaussian mapping module enabled.
Tracking time includes pose estimation and associated online preprocessing, while mapping time includes Gaussian primitives initialization and optimization.
Per-frame times were computed from the accumulated execution time of each stage divided by the total number of processed frames, including non-keyframes.
GPU execution was synchronized for timing, and results were averaged over three runs per sequence, as reported in Table~\ref{tab:stady4d_runtime}.

\begin{table}[t]
\centering
\scriptsize
\caption{Average per-frame tracking and mapping times across three StaDy4D sequences using dynamic RGB inputs.}
\label{tab:stady4d_runtime}

\setlength{\tabcolsep}{3pt}
\begin{tabular}{l | cc}
\toprule
\textbf{Methods}
& \textbf{Tracking}$\downarrow$ [ms/frame]
& \textbf{Mapping}$\downarrow$ [ms/frame] \\
\midrule

MonoGS (2024)~\cite{monogs}
& 579 & \textbf{1134} \\

WildGS-SLAM (2025)~\cite{wildgs}
& 372 & 3587 \\

DROID-W (2026)~\cite{droidw}
& \textbf{249} & 1661 \\

\textbf{DPGS-SLAM (Ours)}
& 510 & 1205 \\

\bottomrule
\end{tabular}
\end{table}

As shown in Table~\ref{tab:stady4d_runtime}, DPGS-SLAM achieved an average tracking time of 510\,ms/frame and a mapping time of 1205\,ms/frame. Its mapping time was comparable to MonoGS and lower than WildGS-SLAM and DROID-W, while tracking remained slower than the latter two methods.

The main tracking overhead arose from iterative pose optimization (292.1\,ms/frame). Motion mask generation required 179.5\,ms/frame, including optical flow estimation (141.8\,ms/frame), OCLR (6.2\,ms/frame), and SAM2 (30.0\,ms/frame). However, parallel execution reduced the blocking time imposed by motion mask generation on tracking to 50.5\,ms/frame. The total tracking time also included data loading, I/O, and other processing overhead.
Diffusion model loading incurred a one-time cost of 5790\,ms, while inference took an average of 390\,ms per selected keyframe. As background completion runs asynchronously, it does not block tracking or observed-map optimization, limiting its impact on processing latency.

\subsection{Ablation Study}

\subsubsection{Tracking Ablation}

To assess the contributions of motion mask generation and geometric constraints to camera tracking, we evaluated three ablated configurations on the four dynamic TUM RGB-D sequences.
\textbf{W/o motion masks} disables both OCLR and SAM2, allowing all valid pixels to participate in tracking and mapping.
\textbf{OCLR only} generates motion masks using OCLR for every frame, without SAM2-based refinement or temporal propagation.
\textbf{W/o geometric loss} removes the geometric term from~\eqref{eq:lossTrack}, retaining only the photometric loss for camera pose optimization.
All other applicable settings remained unchanged.
Table~\ref{tab:ablation_tracking} reports the ATE RMSE for each configuration.

As shown in Table~\ref{tab:ablation_tracking}, the full DPGS-SLAM achieved the lowest average ATE RMSE of 1.2\,cm. Without motion masks, the error increased to 3.1\,cm, while using OCLR alone reduced it to 2.4\,cm, demonstrating the benefit of SAM2-based refinement and temporal propagation. Removing the geometric loss increased the error to 2.3\,cm, confirming the contribution of predicted-depth constraints to camera tracking. These results indicate that accurate motion masking and photometric--geometric optimization jointly support the tracking performance.

\begin{table}[t]
\centering
\scriptsize
\caption{Tracking ablation results on four dynamic TUM RGB-D sequences (ATE RMSE $\downarrow$ [cm]).}
\label{tab:ablation_tracking}

\setlength{\tabcolsep}{3pt}
\begin{tabular}{p{26mm} | cccc | c}
\toprule
\textbf{Configuration}
& \textbf{wk.xyz}
& \textbf{wk.hf}
& \textbf{wk.st}
& \textbf{st.hf}
& \textbf{Avg.} \\
\midrule

W/o motion masks
& 2.6 & 3.4 & 2.1 & 4.2 & 3.1 \\

OCLR only
& 2.0 & 3.1 & 1.3 & 3.0 & 2.4 \\

W/o geometric loss
& 2.3 & 3.1 & 1.1 & 2.9 & 2.3 \\

\textbf{Full DPGS-SLAM}
& \textbf{1.1} & \textbf{1.5} & \textbf{0.4} & \textbf{1.8} & \textbf{1.2} \\

\bottomrule
\end{tabular}

\par\vspace{3pt}

\begin{minipage}{\columnwidth}
\scriptsize
\raggedright
\textit{Notation:}
ATE RMSE is reported in centimeters.
Avg. denotes the mean across the four sequences.
\end{minipage}

\end{table}

\subsubsection{Static-Background Reconstruction Ablation}
\label{sec:static_background_ablation}
We compared five mapping configurations using the same tracking module, depth predictor, keyframe selection policy, and Gaussian-parameter optimization settings.
\textbf{Hard mask} excludes pixels within detected dynamic regions from Gaussian primitives initialization and mapping supervision.
\textbf{Soft mask} allows Gaussian primitives initialization from these pixels and assigns their mapping residuals a weight of $0.1$, while valid pixels outside the masked regions retain unit weight.
\textbf{Adaptive weighting} incorporates WildGS-style~\cite{wildgs} learned per-pixel uncertainty weighting into the same mapping backend.
These three variants omit background completion.
\textbf{W/o replacement} retains background completion, provenance-based isolation of generated primitives, and independent initialization of observation-supported primitives, but disables deletion of generated Gaussian primitives.
\textbf{Full DPGS-SLAM} enables all proposed components.

Table~\ref{tab:ablation_stady4d} compares PSNR, SSIM, and LPIPS inside and outside undilated ground-truth dynamic-object masks shared across methods, using paired static RGB references. The masks were derived from paired dynamic and static ground-truth depth maps by identifying pixels where dynamic objects occlude static surfaces. PSNR was computed from regional RGB MSE, while spatial SSIM and AlexNet (2017)~\cite{alexnet} -based LPIPS maps were computed on full images and averaged within each region. Results were averaged over frames within each sequence and then equally across the three StaDy4D sequences. Inside the masks, Full DPGS-SLAM improved PSNR by $9.9$\,dB and SSIM by $32.2$ percentage points over adaptive weighting, while reducing LPIPS ($\times 100$) from $51.9$ to $36.3$. Compared with w/o replacement, it improved PSNR by $5.6$\,dB and SSIM by $17.5$ percentage points, while reducing LPIPS ($\times 100$) from $43.5$ to $36.3$. Improvements outside the masks are comparatively limited, indicating that the reconstruction gains are primarily concentrated in regions affected by dynamic occlusions. The comparison with w/o replacement further supports the effectiveness of removing eligible generated Gaussian primitives when real observations become available.

Fig.~\ref{fig:ablation_qualitative} presents qualitative comparisons on \textit{wandering} (WildGS-SLAM iPhone dataset) and \textit{freiburg3\_sitting\_halfsphere} (TUM RGB-D dataset), representing insufficient background observations caused by rapid viewpoint changes and human occlusion, and prolonged occlusion by a seated person, respectively.
In both sequences, hard and soft masking produced coarse, blurred backgrounds due to insufficient valid observations to constrain Gaussian refinement, while adaptive weighting retained human-shaped ghosting from residual dynamic observations. Background completion provides provisional representations in these insufficiently observed regions. In \textit{wandering}, disabling replacement allowed diffusion-generated Gaussian primitives to obscure parts of the background subsequently reconstructed from real observations. Full DPGS-SLAM mitigates this issue by removing eligible generated primitives and independently initializing observation-supported Gaussian primitives.

\begin{table}[t]
\centering
\scriptsize
\caption{Ablation results inside and outside dynamic-object masks on three StaDy4D sequences.}
\label{tab:ablation_stady4d}

\setlength{\tabcolsep}{3 pt}
\begin{tabular}{l | ccc | ccc}
\toprule
\multirow{2}{*}{\textbf{Configuration}}
& \multicolumn{3}{c|}{\textbf{Inside mask}}
& \multicolumn{3}{c}{\textbf{Outside mask}} \\
\cmidrule(lr){2-4} \cmidrule(lr){5-7}
& \textbf{\shortstack{PSNR$\uparrow$\\{[dB]}}}
& \textbf{\shortstack{SSIM$\uparrow$\\{[\%]}}}
& \textbf{\shortstack{LPIPS$\downarrow$\\{($\times 100$)}}}
& \textbf{\shortstack{PSNR$\uparrow$\\{[dB]}}}
& \textbf{\shortstack{SSIM$\uparrow$\\{[\%]}}}
& \textbf{\shortstack{LPIPS$\downarrow$\\{($\times 100$)}}} \\
\midrule

Hard mask
& 11.5 & 43.7 & 50.5
& 25.2 & 80.2 & 23.6 \\

Soft mask ($0.1$)
& 13.9 & 45.4 & 51.0
& 25.0 & 81.4 & 22.7 \\

Adaptive weighting
& 12.2 & 41.5 & 51.9
& 25.3 & 80.1 & 22.6 \\

W/o replacement
& 16.5 & 56.2 & 43.5
& 25.9 & 82.7 & 22.0 \\

\textbf{Full DPGS-SLAM}
& \textbf{22.1} & \textbf{73.7} & \textbf{36.3}
& \textbf{26.1} & \textbf{84.8} & \textbf{21.4} \\

\bottomrule
\end{tabular}

\par\vspace{3pt}

\begin{minipage}{\columnwidth}
\scriptsize
\raggedright
\textit{Notation:}
Entries are averaged over frames within each sequence and then equally across sequences.
SSIM and LPIPS are multiplied by 100.
\end{minipage}

\end{table}

\begin{figure*}[t]
\centering
\scriptsize
\setlength{\tabcolsep}{1pt}

\begin{tabular}{@{}cccccc@{}}
\textbf{Input RGB}
& \textbf{Hard mask}
& \textbf{Soft mask ($0.1$)}
& \textbf{Adaptive weighting}
& \textbf{W/o replacement}
& \textbf{Full DPGS-SLAM} \\[2pt]

\includegraphics[width=0.16\textwidth]{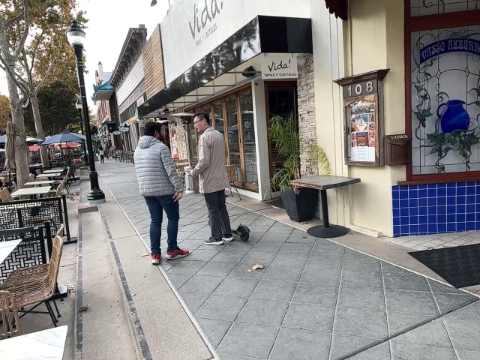} &
\includegraphics[width=0.16\textwidth]{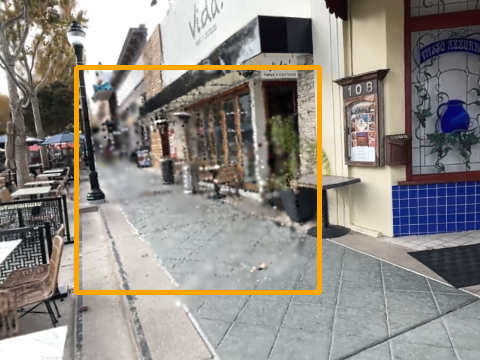} &
\includegraphics[width=0.16\textwidth]{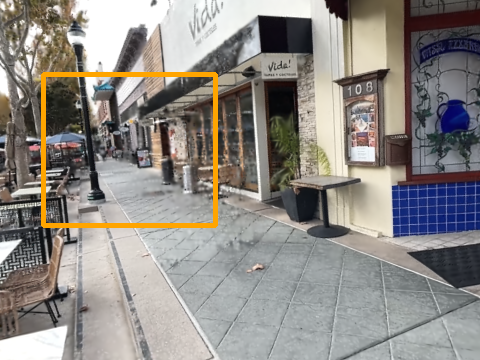} &
\includegraphics[width=0.16\textwidth]{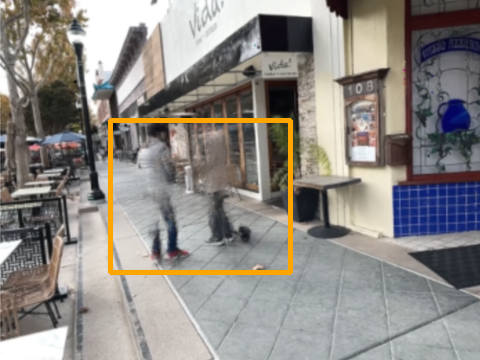} &
\includegraphics[width=0.16\textwidth]{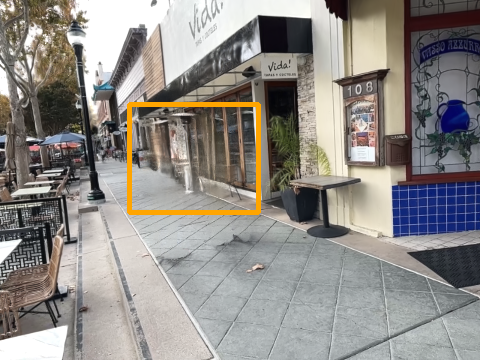} &
\includegraphics[width=0.16\textwidth]{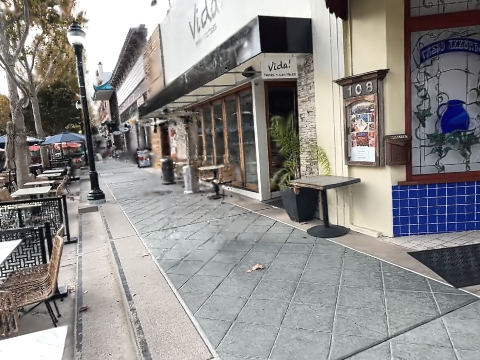}
\\[2pt]

\includegraphics[width=0.16\textwidth]{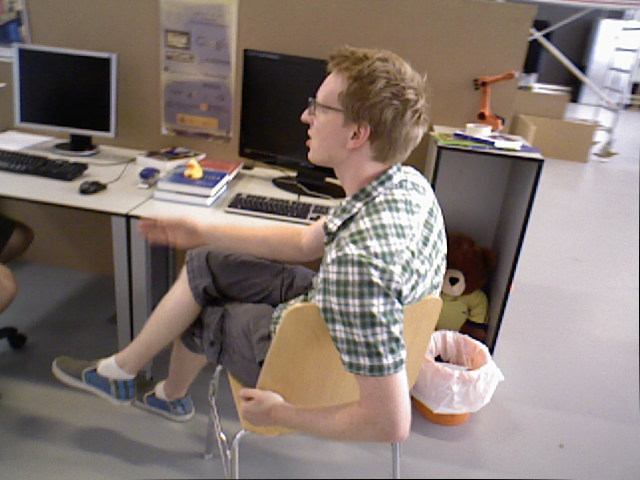} &
\includegraphics[width=0.16\textwidth]{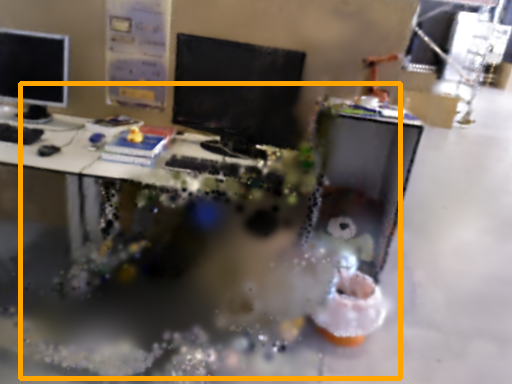} &
\includegraphics[width=0.16\textwidth]{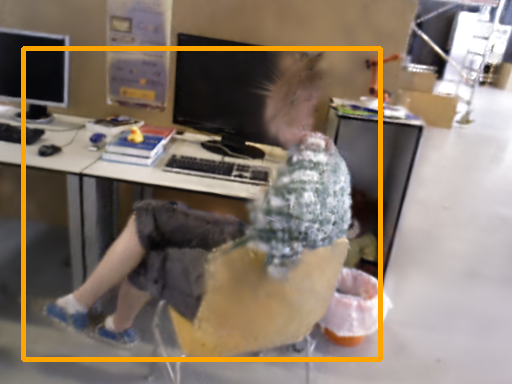} &
\includegraphics[width=0.16\textwidth]{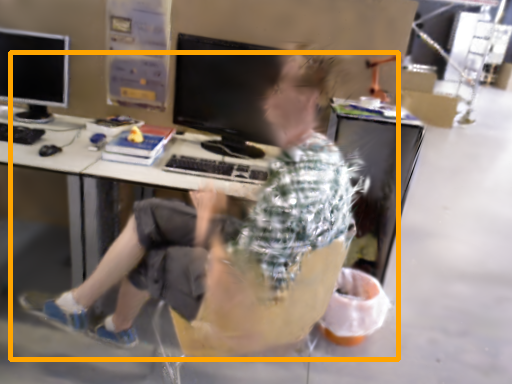} &
\includegraphics[width=0.16\textwidth]{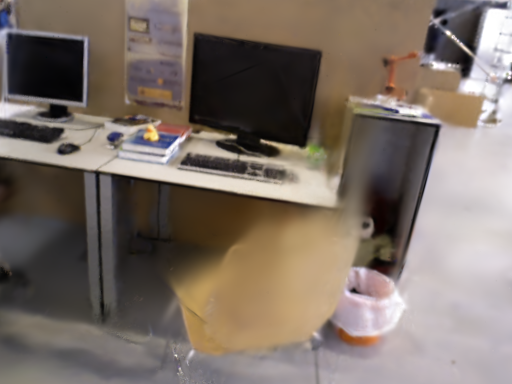} &
\includegraphics[width=0.16\textwidth]{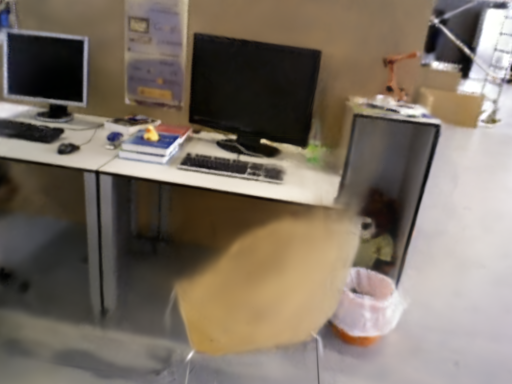}
\\
\end{tabular}

\caption{
Qualitative ablation comparison on \textit{wandering} (WildGS-SLAM iPhone dataset, top) and \textit{freiburg3\_sitting\_halfsphere} (TUM RGB-D dataset, bottom).
The first column shows the input RGB frames, followed by renderings from five configurations. 
The sequences illustrate background reconstruction under limited observations and prolonged occlusion, respectively, with orange boxes highlighting blurring, ghosting, and background occlusion artifacts.
}
\label{fig:ablation_qualitative}
\end{figure*}

\subsubsection{Geometric Accuracy Evaluation}
\label{sec:geometric_accuracy}

To evaluate geometric errors introduced by background completion and depth prediction, we compared Metric3D v2~\cite{metric3d} predictions from diffusion-completed and paired static ground-truth RGB images against ground-truth depth at completion keyframes. We further evaluated rendered depth from three mapping configurations: hard mask, w/o replacement, and Full DPGS-SLAM, as defined in Section~\ref{sec:static_background_ablation}.

For each Gaussian map, opacity-normalized expected camera-axis depth was rendered at ground-truth viewpoints and evaluated after global trajectory alignment. We computed a coverage-penalized variant of AbsRel (absolute relative error) (2014)~\cite{absrel}  separately inside and outside the undilated ground-truth dynamic-object masks, using all pixels with ground-truth depth in $[0.1,80)\,\mathrm{m}$. Missing, nonfinite, or nonpositive predictions, as well as rendered predictions with accumulated opacity below $0.5$, were set to zero, contributing a relative error of $1$ per pixel. Scores were averaged over frames within each sequence and then equally across the three sequences (Table~\ref{tab:geometric_accuracy}).

As shown in Table~\ref{tab:geometric_accuracy}, Metric3D v2 achieved comparable depth accuracy on diffusion-completed and static ground-truth RGB images, with masked-region AbsRel values of 0.053 and 0.049, respectively. This indicates limited additional depth error from background completion. For Gaussian map rendering, hard masking produced a substantially higher AbsRel of 0.523 inside the masks, while w/o replacement and Full DPGS-SLAM reduced it to 0.078 and 0.073, respectively. The improvement reflects more accurate and complete reconstruction under the missing-depth penalty. Observation-based replacement provided a further reduction in masked-region error, although the difference was relatively small. Outside the masks, all three mapping configurations achieved similar accuracy (0.086--0.089), indicating that the geometric benefits are primarily concentrated in regions affected by dynamic occlusions.

\begin{table}[t]
\centering
\scriptsize
\caption{Geometric accuracy on three StaDy4D sequences (coverage-penalized AbsRel $\downarrow$).}
\label{tab:geometric_accuracy}

\setlength{\tabcolsep}{3pt}
\begin{tabular}{p{42mm} | cc}
\toprule
\textbf{Configuration}
& \textbf{Inside mask}
& \textbf{Outside mask} \\
\midrule

\multicolumn{3}{l}{\textit{Depth prediction}} \\
\cmidrule(lr){1-3}

Metric3D v2 (static GT RGB)
& \textbf{0.049} & 0.053 \\

Metric3D v2 (diffusion-completed RGB)
& 0.053 & \textbf{0.051} \\

\midrule

\multicolumn{3}{l}{\textit{Gaussian map rendering}} \\
\cmidrule(lr){1-3}

Hard mask
& 0.523 & 0.087 \\

W/o replacement
& 0.078 & \textbf{0.086} \\

\textbf{Full DPGS-SLAM}
& \textbf{0.073} & 0.089 \\

\bottomrule
\end{tabular}
\end{table}

\subsection{General Discussion and Limitations}
\label{sec:limitations}
The proposed system DPGS-SLAM relies only on monocular RGB input. The diffusion inpainting model was fine-tuned offline using LoRA~\cite{lora}, without additional scene-specific fine-tuning on the evaluated sequences. Metric3D v2~\cite{metric3d} provides predicted depth for tracking, mapping, and provisional Gaussian initialization. Periodic optical-flow-based motion discovery may introduce a brief delay in detecting newly entering moving objects, allowing temporary dynamic interference. Such interference may be mitigated by subsequent motion detection and incremental map updates.

Provenance-aware mapping excludes generated Gaussian primitives from camera-pose estimation and observed-map optimization, and selectively deletes them when valid static observations become available.
However, persistently unobserved regions may retain inaccurate generated appearance or geometry. Asynchronous background completion also incurs computational overhead and does not guarantee real-time performance.

\section{Conclusions}
\label{sec:con}

We present DPGS-SLAM, a monocular RGB 3D Gaussian splatting SLAM system for dynamic environments that provides provisional representations of static backgrounds with insufficient support from real RGB observations. The system combines periodic optical-flow-based motion discovery with SAM2 to obtain temporally consistent motion masks.

For map reconstruction, diffusion-based inpainting enables asynchronous background completion, while provenance-aware Gaussian mapping integrates the generated content as provisional Gaussian primitives.
Generated primitives are excluded from camera-pose estimation and observed-map optimization. When valid static observations become available, eligible generated primitives are removed using a depth-aware projective rule, while observation-supported Gaussian primitives are independently initialized.
Experimental results demonstrated competitive camera tracking, with average ATE RMSE values of 1.2\,cm on TUM RGB-D and 2.7\,cm on BONN RGB-D.
On the three evaluated StaDy4D sequences, DPGS-SLAM improved average PSNR by 4.60\,dB over WildGS-SLAM, while ablation studies further support the benefit of observation-based replacement for static-background reconstruction.
Future work will explore tighter integration of dynamic object modeling and unified optimization formulations to further extend the applicability of the proposed framework.
\section*{Acknowledgment}
The authors acknowledge the use of OpenAI's ChatGPT to assist with grammar checking and language editing. The authors take full responsibility for the scientific content, study design, data analysis, and interpretations presented in this paper.

\bibliographystyle{IEEEtran}
\bibliography{references}

\end{document}